%% file: main.tex
\documentclass[5p,times]{elsarticle}
\journal{Future Generation Computer Systems}

\usepackage{textcomp}
\usepackage{hyperref}
\usepackage{lipsum}

\usepackage{hyperref} 
\usepackage{subcaption}
\usepackage{multirow}
\usepackage{multicol}

\begin{document}

\title{Towards Automatic Model Specialization \\
for Edge Video Analytics}

\author[1,2]{Daniel~Rivas\corref{cor1}}
\ead{daniel.rivas@bsc.es}

\author[3]{Francesc~Guim}
\ead{francesc.guim@intel.com}

\author[1]{Jord\`a Polo}
\ead{jorda.polo@bsc.es}

\author[4]{Pubudu M.~Silva}
\ead{pubudu.silva@intel.com}

\author[1]{Josep Ll.~Berral}
\ead{josep.berral@bsc.es}

\author[1]{David~Carrera}
\ead{david.carrera@bsc.es}

\cortext[cor1]{Corresponding author}

\affiliation[1]{organization={Barcelona Supercomputing Center (BSC)},
                addressline={C. Jordi Girona, 1-3},
                postcode={08034},
                city={Barcelona},
                country={Spain}}
                
\affiliation[2]{organization={Universitat Polit\`ecnica de Catalunya (UPC)},
                addressline={Campus Nord, Edif. D6, C. Jordi Girona, 1-3},
                postcode={08034},
                city={Barcelona},
                country={Spain}}
                
\affiliation[3]{organization={Intel Corporation Iberia},
                city={Barcelona},
                country={Spain}}
                
\affiliation[4]{organization={Intel Corporation},
                city={Hillsboro, OR},
                country={USA}}

\begin{abstract}
Judging by popular and generic computer vision challenges, such as the ILSVRC (ImageNet Large Scale Visual Recognition Challenge) or PASCAL VOC, neural networks have proven to be exceptionally accurate in some tasks, surpassing even human-level accuracy. However, state-of-the-art accuracy often comes at a high computational price, requiring equally state-of-the-art and high-end hardware acceleration to achieve anything near real-time performance. At the same time, use cases such as smart cities or autonomous vehicles require an automated analysis of images from fixed cameras in real-time. Due to the vast and constant network bandwidth these streams would generate, we cannot rely on offloading compute to the omnipresent and omnipotent cloud. Consequently, a distributed edge cloud should take over and process images locally. However, by nature, the edge cloud is resource-constrained, which puts a limit on the computational complexity of the models executed in the edge. Nonetheless, there is a need for a meeting point between the edge cloud and accurate real-time video analytics. 
One solution is to use methods to specialize lightweight models on a per-camera basis but it quickly becomes unfeasible as the number of cameras grows unless the process is fully automated.
In this paper, we present and evaluate COVA (Contextually Optimized Video Analytics), a framework to assist in the automatic specialization of models for video analytics in edge cloud cameras. COVA aims to automatically improve the accuracy of lightweight models through the automatic specialization of models. Moreover, we discuss and analyze each step involved in the process to understand the different trade-offs that each one entails. Additionally, we show how the sole assumption of static cameras allows us to make a series of considerations that greatly simplify the scope of the problem and, in turn, enables COVA to successfully specialize models using traditional computer vision techniques specifically chosen for the task.
Through COVA, we show that complex neural networks, i.e., those able to generalize well, can be effectively used as \textit{teachers} to annotate datasets for the specialization of lightweight neural networks and adapt them to the specific context in which they will be deployed. This allows us to tailor models to increase their accuracy while keeping their computational cost constant and do so without any human interaction. Results show that COVA can automatically improve pre-trained models by an average of 21\% on the different scenes of the VIRAT dataset.
\end{abstract}

\maketitle

% 1.
\input{introduction}
\input{related_work}
\input{problem_breakdown}
\input{framework}

% 5.
\input{methodology}

% 6.
\input{results}

% 7.
\input{conclusions}

\bibliography{references.bib}

% that's all folks
\end{document}

%% file: introduction.tex
\section{Introduction}
\label{sec:introduction}

The edge cloud aims to provide compute and storage resources closer to where the data is produced to filter the amount of data crossing the backhaul and alleviate it. Nevertheless, a geographically distributed edge cloud entails two characteristics whose implications are often undermined. First, different physical locations (at which resources are installed) involve different sets of constraints that ultimately impact the type and size of the resources that can be installed~\cite{rivas2021performance}. For example, a street cabinet might be limited to a handful of server-grade nodes, while a solar-powered node in a street pole might have to settle for an IoT gate with a low-power processor. Regularly, resources become scarcer the closer we are to the edge of the network. Consequently, the more challenging yet more critical it becomes to leverage those resources efficiently. 

Second, the workloads (i.e., services) and the distribution of the incoming requests are tightly coupled to the geographical area the edge location is serving. This relation has long been experienced and exploited in traditional CDNs (Content Delivery Networks). Several factors, such as the time of the day or local trends, shape users' access patterns to the different internet locations daily. CDNs leverage this information aiming to predict what users will access next and cache it near them. Together, the combination of scarce resources, workloads coupled to a geographical location, and any other outside-world data that might affect the workload makes the context of the edge deployment up.

Video analytics, powered by all kinds of DNNs (Deep Neural Networks), is already one of the prominent use cases being executed at the edge~\cite{zhou2019edge}. At the same time, the edge has been appointed as a required actor and the primary accelerator for large-scale video analytics~\cite{ananthanarayanan2017real,bilal2017edge}. All over the world, cameras generate constant data streams whose contents remain largely disregarded unless a specific event triggers their review. Some of these events require immediate action (e.g., alert security if someone breaks in a store or open a barrier after a car approaches it upon checking credentials), while others involve bulk analysis (e.g., crowd counting at a fair or car counting for traffic analysis in smart cities). However, in both cases, action and information are taken at a local level. Therefore, images can be processed in the edge while only a fraction of the data (the results) must be sent back to centralized locations to be registered.

\begin{figure}[!ht]
    \centering
    \includegraphics[width=\linewidth]{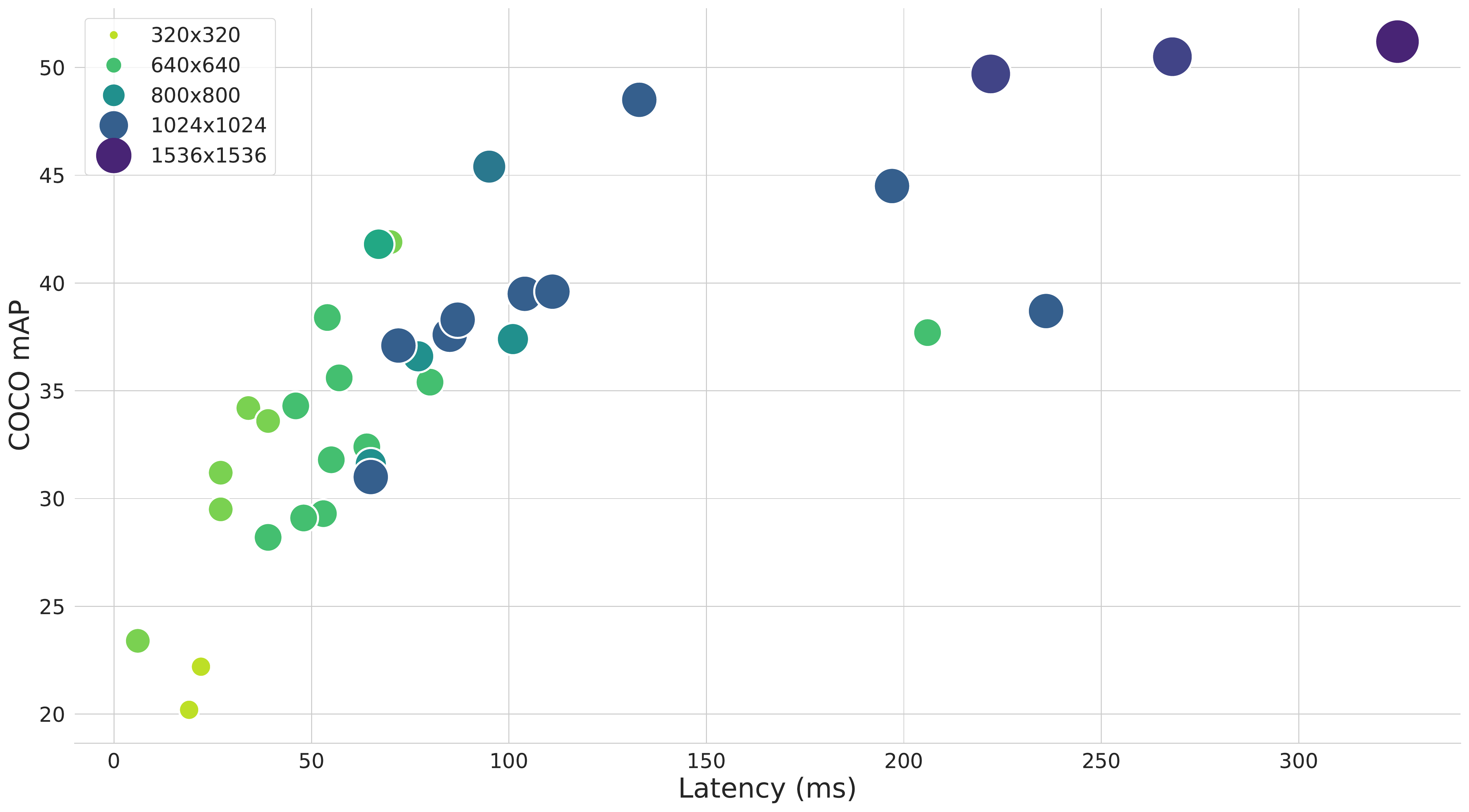}
    \caption{Speed versus mAP (mean Average Precision) on the COCO for different object detection pre-trained models~\cite{repo:tf_modelzoo}. Bubble size represents the input size of the model.}
    \label{fig:speed_vs_map}
\end{figure}

Moreover, there is a direct relationship between the accuracy of a trained neural network and the size of the network (i.e., number of trainable parameters). Unfortunately, bigger networks tend to have higher computational complexity than simpler ones, as depicted in Figure~\ref{fig:speed_vs_map}. Ultimately, resource constraints on edge nodes set a limit to the size of the neural networks that execute and, thus, also to the accuracy they can achieve. State-of-the-art accuracy is usually out of the picture for the so-called \textit{edge DNNs} and is only within reach of bigger neural networks. The difference between these two may imply orders of magnitude more trainable parameters, and the gap is only increasing with time~\cite{huang2017speed}. At the same time, hardware acceleration becomes a must to provide anything near real-time performance. Altogether, these limitations pose a challenge for the edge cloud to postulate itself as a serious alternative to current data centers for video analytics unless the hardware installed is upgraded or the models deployed are greatly optimized.

One of the main reasons for bigger models to achieve higher accuracy is that the extra trainable parameters can be used to learn more robust high-level features that allow the model to generalize its predictions in new environments successfully. Simpler models may face difficulties in capturing the relationship between the input examples (i.e., images) and the target values (i.e., predictions), which is known as \textit{underfitting}. However, it is essential to emphasize that simple or big are adjectives relative to the scope of the problem, and neural networks should be dimensioned according to the level of generality and complexity required for the task at hand. Otherwise, the \textit{underfitting} we experience whenever we expect too much from a model may turn into \textit{overfitting} when the network is big enough to memorize the training examples, which will translate into poor generalization on unseen examples.

Several popular challenges evaluate neural networks on specific datasets to tell how well a neural network architecture might perform. For example, ResNet152 scores 94.5\% Top-1 accuracy on the ImageNet~\cite{imagenet} dataset, while a smaller edge model like a MobileNetV2 is limited to \textit{only} 71\% Top-1 accuracy on the same challenge~\cite{bianco2018benchmark}. However, challenges like ImgeNet or COCO (Common Objects in Context)~\cite{lin2014microsoft} are notoriously generalists, and models are required to distinguish a car from a toothbrush while also telling a \textit{water snake} and a \textit{vine snake} apart. Such diversity of objects is rarely captured by static cameras deployed to the edge. Just as a camera in the wild will hardly see many trucks, a camera controlling traffic at an urban intersection will not capture many wild animals. By narrowing down the diversity of objects to detect, simpler neural networks become available to be executed in the edge in real-time.

Nevertheless, the loss of generality brought about by the specialization of the model also implies that models are now potentially tight to a camera's context. Where we previously had a single model for the analysis of many cameras, we now potentially need one model per camera, each requiring its very own training process with properly annotated training examples. One training per camera quickly becomes unrealistic as the number of cameras grows, limiting the potential benefits of model specialization unless the process can be optimized and completely automated. 

% In this paper, we present and evaluate COVA (Contextually Optimized Video Analytics), a framework to optimize edge video analytics workloads by taking advantage of the context of edge deployments and specialize the application for such context.
This paper presents COVA (Contextually Optimized Video Analytics), a framework to automate the specialization of models for real-time edge video analytics. Through it, COVA funnels the generality of big neural networks with state-of-the-art accuracy into specialized edge neural networks, resulting in models with high accuracy but on a narrowed scope. Our work is based on a simple observation: \textit{static cameras do not move, but objects do}. At the same time, in the context of edge video analytics, we can safely assume that moving objects are equivalent to objects of interest, which allows us to leverage computer vision techniques to simplify the task. Moreover, we define the context of a camera as the set of invariant properties associated with its deployment location. For example, height or focal distance determine the perspective from which objects are seen. Similarly, the background and the environment (e.g., urban area, indoor, or nature) determine the type of objects that we can expect the camera to capture. Once the context of a camera is defined, automating the specialization of models can be vastly simplified. 

%\textcolor{red}{COVA Once a camera is deployed, COVA captures images and selects the most representatives. 
%Then, these images are annotated by a \textit{ground-truth model} executed in a remote data center. The ground-truth model acts as an oracle or \textit{teacher} whose predictions are taken as ground-truths during the training of the edge model, working as \textit{student}. By assuming that images are only from static cameras, the framework can use traditional computer vision techniques to determine the representativeness of each frame. At the same time, these same techniques are also used to boost the accuracy of the ground-truth model and improve the quality of the resulting edge model.}

The contributions of this paper are two-fold.
On one hand, we present and evaluate COVA (Contextually Optimized Video Analytics), a framework to assist in the rapid deployment of tailored edge models for real-time video analytics. The pipeline in COVA is conceived to be highly modular and easily extensible to allow for different types of contextual analysis of the scenes. It supports TensorFlow~\cite{tensorflow} (training and inference) and OpenVINO~\cite{openvino} (inference only) and can run either local or on \textit{Amazon Web Services} (AWS) and orchestrated from the edge. Moreover, source code is made publicly available~\cite{edge_repo}.
On another hand, we present an extensive exploration and review of the different steps involved in the process of automating the specialization of models, including considerations and findings to be taken into account in case some of the pipeline stages are to be exchanged or extended.

%involved in the process of automating the specialization of models as well as our findings throughout its  while as well as the different considerations made throughout the   accuracy of edge models without any extra compute cost via automatic model specialization. We explore and evaluate several considerations to automate the specialization of models for edge video analytics effectively. Moreover, we show how the sole assumption of static cameras allows us to simplify the scope of the problem and propose a series of methods that would fail if applied to a broader problem.

The rest of the paper is organized as follows: 
Section~\ref{sec:related_work} revises the previous work and discusses the gaps that our work aims to fill. 
Section~\ref{sec:problem_breakdown} breaks the problem of automating the specialization of models down, and provides the background to understand the challenges behind deploying accurate neural network models to the edge. %and the considerations taken into account while developing COVA. 
Section~\ref{sec:framework} describes the COVA framework and its components.
Section~\ref{sec:methodology} describes the methodology followed throughout the evaluation. 
Section~\ref{sec:results} presents the experimental results and their analysis. 
% Section~\ref{sec:discussion} discusses and explores ways to extend the automatic contextual analysis done by COVA to specialize models.
Finally, Section~\ref{sec:conclusions} presents the conclusions of our work, discusses the results, and introduces future lines of research opened by the COVA framework.

%% file: related_work.tex
\section{Related Work}
\label{sec:related_work}

\subsection{Edge Video Analytics}
The edge cloud has been previously proposed as a key enabler of video analytics ~\cite{shi2016edge}. In some cases, the edge is considered the \textit{only} realistic approach to meet the latency requirements needed for real-time video analytics ~\cite{ananthanarayanan2017real}, while maximizing QoE by minimizing the cost and flow of data between data centers and users ~\cite{bilal2017edge}. However, the edge cloud presents as many opportunities as new challenges that will have to be properly addressed on any new edge deployment~\cite{shi2016edge, kang2019challenges}.

Within the context of edge video analytics, there have been lots of efforts focused on optimizing the data flow between edge devices and the data center. Authors in ~\cite{ali2018edge} propose an edge infrastructure with the pipeline for deep learning execution distributed across different points in the network (edge, in-transit, and cloud resources). Another optimization point consists of filtering what is sent to the data center. Authors in ~\cite{canel2019scaling} present \textit{FilterForward}, a system aimed to reduce bandwidth consumption and improve computational efficiency on constrained edge nodes by installing lightweight edge filters that backhaul only relevant video frames.

\subsection{Model Specialization}
Model specialization is recurrently used as a mean to optimize the inference cost in scenarios where generality is known to be unnecessary. On the one hand, some approaches leverage specialization over time through online training. For example, authors in~\cite{shen2017fast} identify short-term skews in the class distribution often present in day-to-day videos and exploit it by training models specialized for such distribution online. On the other end of the spectrum, authors in~\cite{cai2019once} propose \textit{Once-for-All}, a neural network architecture that can be repurposed and deployed under diverse architectural configurations but is trained only once.

Moreover, model specialization in edge video analytics is core to various video query systems~\cite{kang2017noscope,hsieh2018focus,kang2019blazeit,hung2018videoedge}. Such systems aim to answer queries about a video by analyzing its contents (e.g., return frames containing instances of a red car between two points in time). Authors in~\cite{kang2017noscope} are able to reduce the cost of running queries against a video by up to three orders of magnitude thanks to \textit{NoScope}, a system that uses inference-optimized model search to find and train the optimal cascade of binary classifiers specialized for the video being queried. Authors in~\cite{hsieh2018focus} propose the use of inexpensive specialized NN at ingest time (i.e., when images are captured) to filter the frames to be considered at query time. Nonetheless, these systems focus on the optimization of queries. Therefore, their main metric to optimize is the latency of a query (involving multiple inferences) and not that of a single inference. Consequently, such systems are subject to a different set of requirements and constraints than real-time edge video analytics.

\subsection{Automatic Training}
Model specialization at large-scale can only be feasible if the process is automated. When the goal is to generate specialized models, previous works have focused on distilling previously acquired knowledge from complex and accurate models (teachers) into simpler ones (students)~\cite{gou2021knowledge}. Through distillation, the student model is trained by trying to mimic the output of the teacher model. On the one hand, distillation can be used to transfer all the knowledge of teacher model into the student to reduce its computational requirements while retaining \textit{most} of the teacher's accuracy on the original dataset~\cite{polino2018model}. On the other hand, distillation can be used to generate specialized models in cases where the problem to be solved by the student is a subset of problem for which the teacher was trained. In such cases, the teacher model works as an oracle that is consulted to predict the labels of the training examples for the student~\cite{mullapudi2019online,kang2017noscope}. In this paper, we focus on the latter to automatically specialize models for the context of each camera deployment. 

Other works have focused on self-supervised learning. Authors in \cite{goyal2021selfsupervised, purushwalkam2020demystifying} demonstrate that such methods are able to outperform traditional methods for supervised training in terms of accuracy. However, while very promising, these methods rely on vast amounts of compute resources and tackle the problem from a completely different baseline than does not fit the requirements and restrictions of the edge cloud.

\subsection{Spatio-Temporal Locality}
When analyzing video feeds from static cameras or any other kind of video where the camera remains still, there is a piece of information that is key to guessing where interesting objects may lie: anything new in the scene Vehicles, persons, birds, or anything moving is subject to be identified. Spatio-temporal information can tell us when something has moved within the range of view, and the region around the moving object can be used as the region of interest to start looking. %This can be easily done with traditional computer vision techniques (background subtraction).
Authors in~\cite{mullapudi2019online} leverage the spatio-temporal locality captured by static cameras to drive the online training of a small specialized model, while authors in ~\cite{wang2018fully} propose a training mechanism in which temporal information is used to train a deep neural network at \textit{instance level} (instead of pixel level) and successfully identify even occluded objects.

% "A static camera will always take images of the same scene unless it is moved, which causes the data from one camera to be highly repetitive. Without sufficient data variability, machine learning models may learn to focus on correlations in the background, leading to poor generalization to novel deployments ~\cite{beery2018recognition}." 
%From: https://ai.googleblog.com/2020/06/leveraging-temporal-context-for-object.html

% From the abstract of ~\cite{beery2018recognition}:
% \begin{quote}
%     It is desirable for detection and classification algorithms to generalize to unfamiliar environments, but suitable benchmarks for quantitatively studying this phenomenon are not yet available. We present a dataset designed to measure recognition generalization to novel environments. The images in our dataset are harvested from twenty camera traps deployed to monitor animal populations. Camera traps are fixed at one location, hence the background changes little across images; the capture is triggered automatically, hence there is no human bias. The challenge is learning recognition in a handful of locations, and generalizing animal detection and classification to new locations where no training data is available. In our experiments \textbf{state-of-the-art algorithms show excellent performance when tested at the same location where they were trained}. However, we find that generalization to new locations is poor, especially for classification systems.
% \end{quote}

Beery et. al take this approach a little bit further with Context R-CNN \cite{beery2020context} and propose a neural network that is divided into two stages: the first one is trained as a traditional R-CNN for object detection whose region proposals are used to train an LSTM. Stage 2 is an attention block that will consider the information of previous frames where detected objects have already appeared. The authors used video feeds from the span of more than one month to train the Context R-CNN. This method is able to correctly identify and detect objects even under challenging environmental conditions such as dense fog. However, authors in \cite{beery2018recognition} make the observation that without sufficient data variability, models learn to focus on correlations in the background, which leads to poor generalization to new deployments. In this paper, we propose to take the previous observation and invert the variable to optimize. Instead of finding methods to generalize to new deployments, we propose to specialize models for each deployment using images from the same static camera. By removing the generalization as the goal, simple and lightweight architectures for object detection become available, which is a requirement to run real-time video analytics in resource-constrained nodes in the edge.

% We take this observation ("state-of-the-art algorithms show excellent performance when tested at the same location where they were trained") and use it to our advantage. 

% In this paper, we analyze the time required to tune lightweight models to perform at least as good as some state-of-the-art algorithms that are able to generalize better.

This paper presents and evaluates COVA, a framework to provide automatic specialization of models for real-time edge video analytics and tools to easily develop custom video analytics applications. Moreover, COVA is envisioned with extensibility in mind and could be used in conjunction with some of the methods mentioned above to further increase accuracy or speed. To the best of our knowledge, no previous work provides the framework and tools that COVA provides.

%% file: problem_breakdown.tex
\section{Problem Breakdown}
\label{sec:problem_breakdown}
Convolutional Neural Networks (CNNs) are able to learn abstract high-level representations of the objects they are trained to detect. The rule of thumb is that assuming proper training, the bigger the network, the better it can learn high-level features to generalize with high accuracy on new data. However, the deployment of generalist CNNs with state-of-the-art accuracy raises three main difficulties:

\begin{itemize}
    \item They require a vast amount of data adequately labeled.
    \item The training phase often takes hundreds, if not thousands, of hours, even with high-end accelerators (such as GPUs or TPUs).
    \item Complex models may provide good accuracy but are not suitable to be executed on resource-constrained edge nodes, while simpler models suitable for the edge achieve lower out-of-the-box accuracy that is often not enough for many use cases.
\end{itemize}

Thanks to automating the retrieval, selection and annotation of the training images (even if it results in an imperfect annotation) and tailoring or specializing lightweight models, we can provide high accuracy models with real-time performance that can be reliably trained within minutes (or less than an hour on CPU-only).

\begin{figure}[h]
     \centering
     \includegraphics[width=\linewidth]{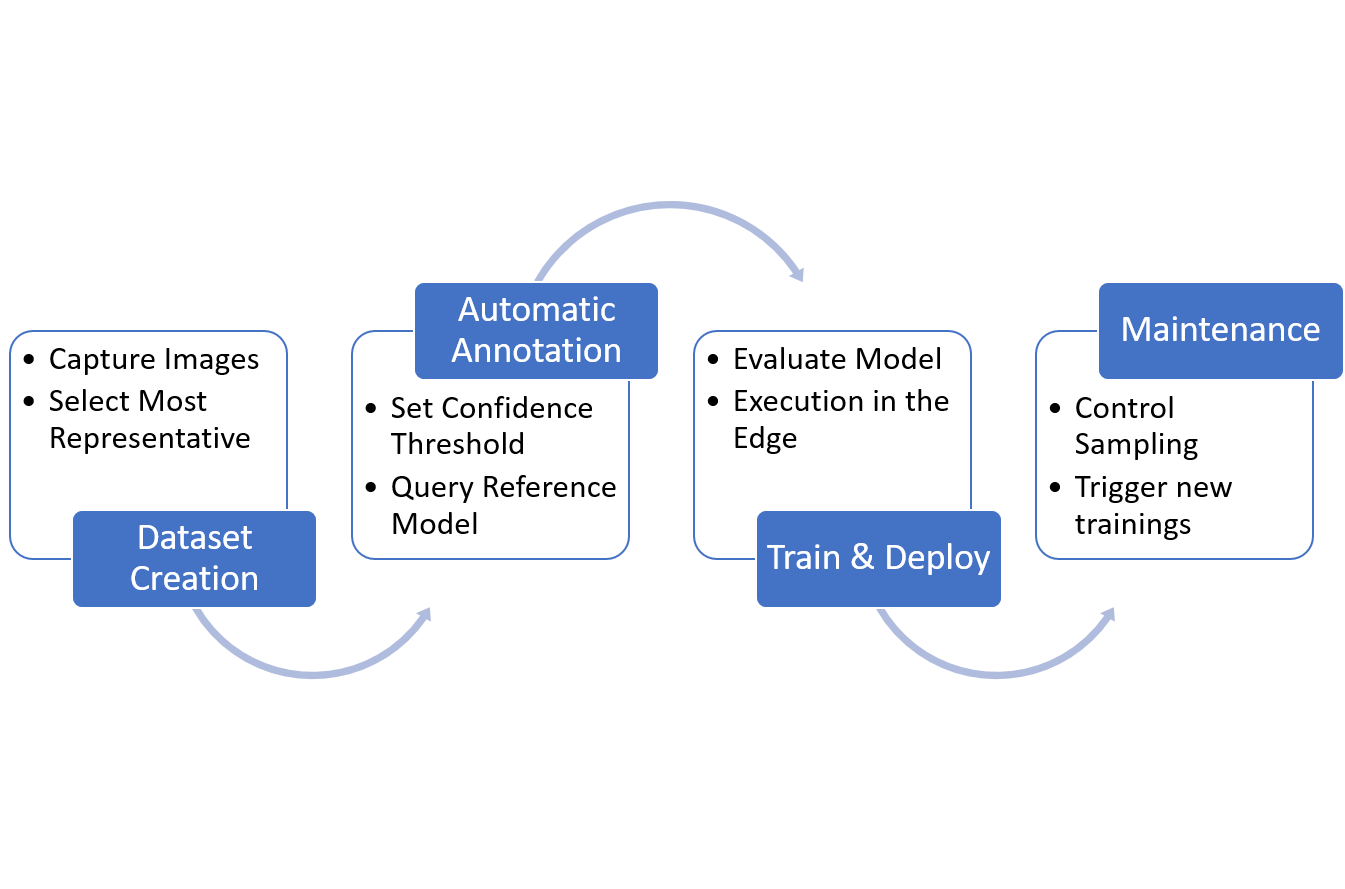}
     \caption{Breakdown of the different steps involved in the automation of specializing models.}
     \label{fig:unsupervised_pipeline}
\end{figure}

The end-to-end process of specializing models is composed of various steps, each with its own set of challenges. Therefore, the automation of the process with satisfactory results requires several considerations to be made. This section breaks the problem of model specialization down into its core parts to analyze its challenges and considerations. Figure~\ref{fig:unsupervised_pipeline} depicts the different steps involved in the process. Once images are captured by the camera, the most representative are selected to be annotated by the ground-truth model and be included in the training dataset. Following, the edge model is trained using the generated dataset. Once deployed, the model's accuracy can be monitored to trigger new training iterations if it falls below a certain threshold.

%\textcolor{red}{We divide these steps into four main building blocks. The first two blocks are the creation of the training dataset and its automatic annotation, in which images from the edge camera are captured, selected, and, finally, annotated. These are what we consider to be the most important blocks, as the final result of the solution will highly depend upon the representativeness of the selected images and the correctness of their annotation. Third, the training or fine-tuning of the edge model using the training dataset created on the previous blocks and its deployment to the edge nodes where the cameras are installed. Finally, the maintenance of the model to monitor its accuracy within accepted intervals of accuracy.}

%Continuous video analytics on the edge usually relies on resource and power constrained nodes were we cannot expect any type of hardware acceleration (although possible, we should not rely on its presence). Therefore, we are forced to optimise every step of the process as much as possible, if the edge is expected to take over these use cases.

% \begin{figure}[!ht]
%     \centering
%     \includegraphics[width=\linewidth]{images/baseline_map.pdf}
%     \caption{Comparison of mAP of different pre-trained models on MS COCO and the VIRAT datasets.}
%     \label{fig:ots_accuracy}
% \end{figure}

\subsection{Static Context from Static Cameras}
Before we delve into the specifics of the problem and the solution we propose, we would like to highlight the main observation that makes everything possible: static cameras do not move. While this observation may seem obvious, it simplifies the problem and the solution to a great extent. 

The training of object detection models requires a large number of diverse training examples. Object instances in the training dataset should appear in all their diversity (e.g., different colors, shapes, or perspectives). This diversity allows the neural network to learn high-level features inherent to the object class and invariant regardless of the specifics of the observed instance. In the end, diversity during training lets the model generalize. At the same time, a static camera is a camera whose position and orientation remain invariable throughout time. That is, it always points at the same scene using the same lens and from the same distance and point of view. Therefore, thanks to assuming static cameras, we are able to define the context of a camera and, then, tailor models for it. %\textcolor{red}{We define the context of a camera as the set of invariant features present in the scene.}

A static camera thus allows the effective optimization of the pipeline by applying simple traditional computer vision techniques. Specifically, we make use of simple techniques for background subtraction with two purposes: (1) select for training and annotation only those frames where there was any meaningful change and (2) let the \textit{ground-truth} model use the image resolution to its full extent by focusing on the parts that changed.

\subsection{Capture (Relevant) Training Images}
Each image annotated by the ground-truth model has a cost associated in terms of network bandwidth and compute power in the data center. However, not all images captured by a camera provide the same amount of new information (if any, at all). Video streams from static cameras tend to have a high degree of \textit{temporal locality} as frame rates between 10 to 30 frames per second are standard for edge cameras. Such frame rates translate to new images being captured every 100ms to as low as 30ms. In most scenarios, nothing new happens within 30 milliseconds, but simply increasing the time between frames does not guarantee that new objects enter the scene. Therefore, we should find the right set of images that will maximize the accuracy of the resultant edge model while minimizing the number of queries to the data center.

Luckily, we can exploit the temporal locality to easily detect changes or quantify similarity between consecutive frames. As the scene's background remains mostly still throughout time, change can be detected by simply comparing frames and set a threshold to decide whether a frame should be considered or not. Relatively simple and inexpensive distance metrics, such as the Mean Squared Error (MSE), can be used to detect meaningful changes between frames~\cite{kang2017noscope}, while other mechanisms, more robust but also more computationally expensive, rely on Siamese neural networks~\cite{daudt2018fully}.

Furthermore, thanks to the same temporal locality, we can also leverage robust motion detection and background subtraction techniques. When working with static cameras, we can consider change to be caused by movement. Therefore, we can identify and isolate the regions of the scene that contain movement, i.e., regions of interest. For such purposes, different approaches with different accuracy-cost trade-offs are available, from those that make use of traditional computer vision techniques~\cite{benezeth2010comparative} to more modern that make use of neural networks~\cite{bouwmans2019deep}.

%\textcolor{red}{To solve this problem, we use motion detection techniques to select only those images containing new objects. This allows us to virtually skip entirely the annotation of those images that would not add new instances to the training examples. Therefore, thanks to motion detection we optimize the computational cost while increasing the representativeness of those images that are selected for annotation.}

COVA, as detailed in Section~\ref{sec:framework_architecture}, makes use of traditional computer vision techniques to model the background and detect and extract the regions containing movement. These regions are regularly smaller than the full scene and can be compared against the same region in previous frames to quantify the level of similarity at a finer grain. Moreover, the extraction of such regions brings an unexpected benefit that, according to our results, boosts the accuracy of the annotations~\cite{rivas2021large}. Nowadays, FullHD video (1920x1080 pixels), if not 4K (3840x2160 pixels), is widely used even among inexpensive cameras available on the market. However, neural networks are rarely deployed with an input resolution matching the camera's resolution, as the computational cost of their execution would become intractable. However, it does not appear to be a limitation on technology but rather a matter of diminishing returns. On the contrary, neural networks do not require large pixel densities to distinguish objects of different classes. An input size of 768x768x3 (as the ground-truth model used throughout the paper) is above average, judging by what can be found in the literature~\cite{repo:tf_modelzoo}, while 300x300x3 pixels is a typical size for edge models (like the one used throughout the paper).

\begin{figure}[h]
     \centering
     \includegraphics[width=.8\linewidth]{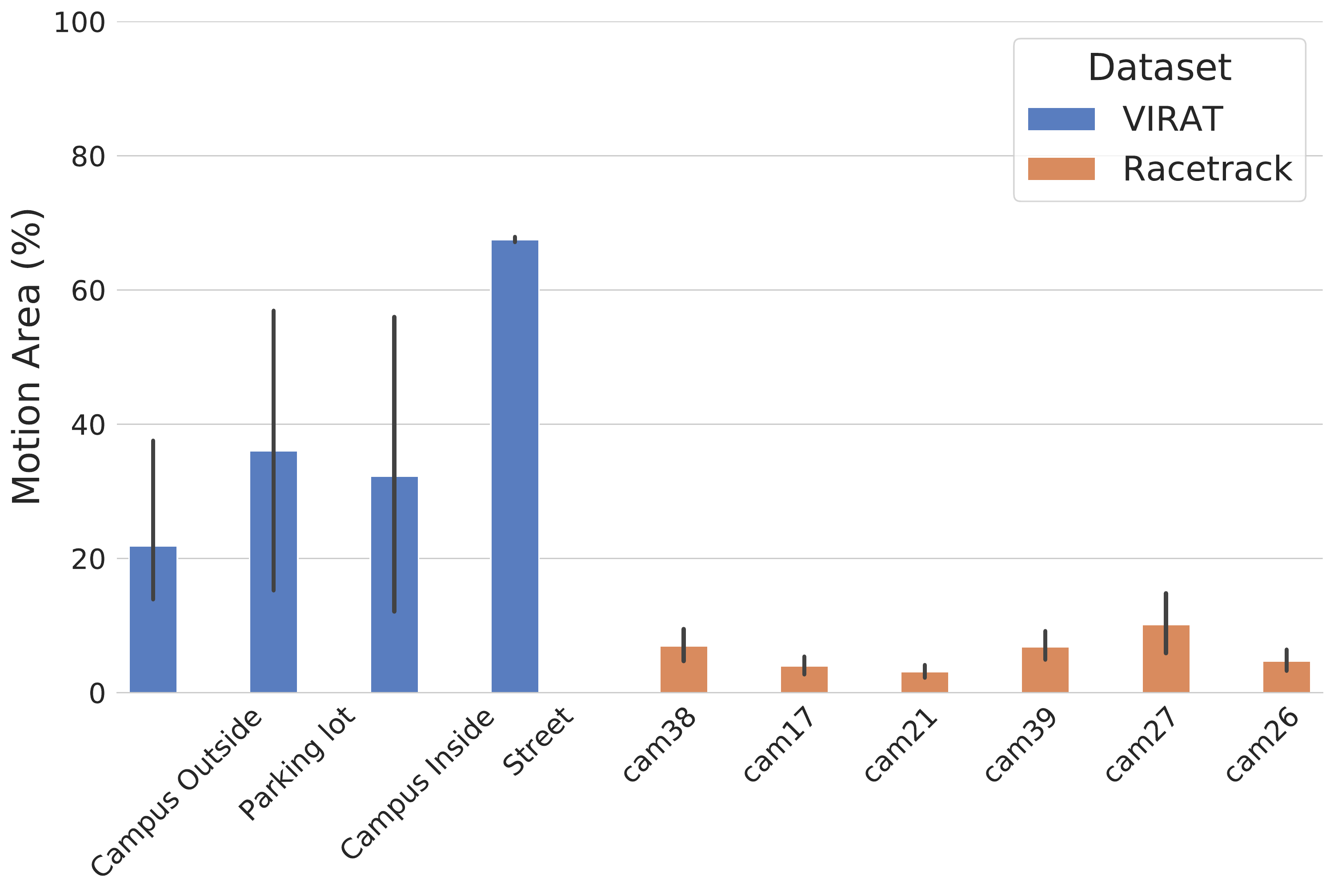}
     \caption{Average percentage of the frame's area containing motion for different scenes. Frames from the Racetrack dataset have much smaller areas, as cars in the track can only drive through the asphalt and always follow the same paths. The VIRAT dataset contains scenes with up to 67\% of the area covered by objects of interest, common in busy urban scenarios with objects appearing from multiple points.}
     \label{fig:area_moving_objects}
\end{figure}

If we consider object detection in the outside world, small input sizes may indeed become a problem. In these scenarios, objects are captured from arbitrarily large distances, and the smaller objects quickly turn into a small set of indistinguishable pixels as the image is resized to the neural network's input size. In such cases, it is worth considering the image to its full extent. The same motion detection techniques we use to optimize the collection of images allow us to do this. Once motion is detected, COVA crops the bounding box containing all frame regions that changed with respect to the background model. Consequently, regions that did not add any information are discarded, and the model can focus on detecting whatever was that moved.
For example, before passing the image to the ground-truth model, we have compressed the information captured by the camera between 3.5 and 14 times for FullHD and 4K images, respectively (or between 23 and 92 times for the edge model). As Figure ~\ref{fig:area_moving_objects} shows, the average percentage of the frame's area containing motion in different scenarios is often a small fraction of the total area. According to the results, these regions represent as little as 3\% of the scene's area.

\subsection{Model Specialization}
To deploy video analytics to multiple cameras (each one with its own context), we could take two main directions. On the one hand, we could run big and complex CNNs capable of generalizing and delivering high accuracy regardless of the cameras' context. This solution could work out-of-the-box if the model has already been trained to detect the same set (or a superset) of object classes as those we want to detect in the edge. Moreover, this solution would require a single training for multiple cameras. However, the reality is that edge deployments do not have the luxury to run such models whose compute requirements are rarely met at the edge. Accuracy comes at the price of high latency and low throughput, often below the minimum required to run real-time use cases. On the other hand, we can use lightweight neural networks but \textit{specialize} them to each specific deployment and let them deliver their full potential on a narrower scope.

The edge cloud is limited on the size of the neural networks that can run. At the same time, this puts a limit on the level of accuracy that a model can achieve with respect to that in the state-of-the-art~\cite{bianco2018benchmark}. Hence, models deployed to the edge may face difficulties extrapolating their knowledge to detect objects in new environments. The specific lighting conditions, points of view, or focal distance of every deployment may impact the ability of small neural networks to extrapolate using previously learned features. One approach to mitigate the effects of lack of generalization from smaller models is called \textit{model specialization}, which implies narrowing the scope of what the network has to learn and, thus, demanding less from it. In the context of edge video analytics, model specialization allows us to adapt a model for the specific context of each camera deployment. The specifics of a context may not be known beforehand but can be easily extracted once the camera has been deployed and begins to capture images. 

Model specialization can be achieved through different means. One option is to train the network from scratch on a specialized dataset. However, the computational cost plus the number of labeled training images required to train a network from scratch on a per-camera basis makes it unfeasible for rapid deployments. Nevertheless, different techniques, such as model distillation~\cite{hinton2015distilling} or transfer learning, allow us to use previously learned knowledge and repurpose it faster while requiring fewer training images. In this paper, we focus on transfer learning to specialize pre-trained models on the specifics of each deployment. The knowledge of a pre-trained generic model with state-of-the-art accuracy, working as \textit{teacher} or \textit{ground-truth model}, is distilled into a specialized training dataset used to train the \textit{student} or \textit{edge model}. %\textcolor{red}{This is a form of model distillation... NO NEW KNOWLEDGE (check notebook).}

Nonetheless, a neural network's predictions are as good as the dataset used to train the network, and transfer learning still requires properly labeled images to specialize models. A proper training dataset is essential to produce an accurate yet small model that can process images in real-time on constrained nodes.

%Figure \ref{fig:ots_accuracy} shows the mean Average Precision of different pre-trained models for the COCO and VIRAT datasets. We can observe how even the least performing model achieves close to 20\% mAP on the COCO dataset, which consists of 80 classes of all kinds in different environments, is still above the mAP achieved by the bigger models on the VIRAT dataset, even when VIRAT contains a subset of pretty common classes like person, car, and bike.

%\subsubsection{Fine-Tuning or Model Specialization}
%Once the training dataset is built, we are ready to start fine-tuning the model that will run on edge nodes. There is no restriction on where this step is executed and may occur in data center or in the very same locations where the models will later run, interchangeably. The specific characteristics of the deployment will determine the location upon the costs of each. For example, fine-tuning a MobileNetV2 feature extractor with an SSD detection head for 1000 epochs with 1000 images requires \~ 30 minutes of CPU time in an Intel Core i5 ****U and a total of **** Joules of energy. The same model, requires just 8 minutes in GPU, while its energy consumption goes up to **** Joules. On the other hand, in this paper we make use of the \textit{Racetrack dataset}, whose images were captured by nodes powered by solar panels with a battery attached. In the morning, there are days when these nodes have their batteries at full capacity and are still consuming less than what the panels can provide. In such cases, any energy not consumed is energy that is wasted.

\subsection{Concept Drift Detection}
\textit{Concept drift} refers to a change in the statistical properties of the variable that a model tries to predict. In the context of edge deployments, concept drift could arise in different ways. For example, a camera in the wild would capture predominantly green backgrounds during spring that turn white during winter due to snow; a camera whose perspective changes substantially after it is relocated; a change in light conditions from day to night; or simply because the user is interested in new classes. A new training should be triggered to adapt the models to the new environment or contextual features in all these cases.

This said, concept drift detection is not trivial. Traditional approaches try to detect concept drift by performing statistical hypothesis testing on multivariate data ~\cite{dries2009adaptive}, while more recent ones explicitly designed for neural networks propose the use of an oracle or teacher model that is consulted intermittently to asses the deployed model's accuracy~\cite{mullapudi2019online} and drive the training rate whenever accuracy falls below a threshold.

Nonetheless, it is important to highlight how, in the context of static cameras, concept drift is highly coupled to the level of specialization of the deployed model. A highly-specialized model trained using only daytime examples may face difficulties in detecting the same objects during nighttime due to lighting variations. On the contrary, a model trained with all-day examples and big enough to retain the sufficient level of generality will not experience a concept drift as day turns to night. Consequently, prediction cascades can help mitigate the problems associated with concept drift without requiring retraining. In prediction cascades, multiple specialized models (either binary~\cite{kang2017noscope} or multi-label classifiers~\cite{wang2017idk}) are sorted by decreasing level of specialization (or increasing level of accuracy and cost) and a model is only consulted if the previous one is not confident enough of its predictions. In the event of a short-term concept drift that cannot be absorbed by the more specialized models, prediction cascades use the more expensive models as fallback to ensure the accuracy of the predictions. However, in the event of a long-term concept drift, the specialized models will be rendered useless and counterproductive, as the cascade will keep consulting them even when they can no longer produce reliable predictions. Therefore, prediction cascades can mitigate the problem but do not solve it. As such, they are a better suited for query systems where the model search can consider the full video stream being queried.

%\textcolor{green}{In this paper, we focus our results on the first training iteration. Therefore, this step is not yet fully automated, but, in our implementation, we use a simple approach to detect concept drift in which the client periodically selects images to be annotated by the ground-truth model and then checks the relative accuracy of the model. If accuracy falls below a certain threshold, a new fine-tuning may be triggered.}

\subsection{Incremental Learning}
The detection of a concept drift irredeemably results in a new training of the model to adapt it to the new environment. This is especially true in ever-changing scenarios, like seasons in a forest, where we have to rely on \textit{incremental learning} to train a model with whatever data is available at the time of the deployment and, as new examples appear, retrain the model using the newly captured examples. However, there is one crucial limitation: \textit{catastrophic forgetting}\cite{french1999catastrophic}, which refers to the tendency of neural networks to forget previously learned features entirely upon learning new ones. The straightforward solution is to train again using the entire dataset, i.e., include the newly captured examples into to the training dataset used during previous iterations. However, this can lead to an oversized dataset that can quickly become intractable for edge nodes, while older examples may not be significant anymore in the current environment.
When the set of classes that are missing is known, another solution is to add generic instances for those classes (e.g., using online image search engines~\cite{li2019rilod} or reusing a subset of the examples used during the training of the ground-truth model or other \textit{students}). Nonetheless, previous work demonstrates that using knowledge distilling techniques~\cite{hinton2015distilling} networks can retain most of the previously learned features, even if trained only on new classes of objects~\cite{kirkpatrick2017overcoming,shmelkov2017incremental}.

%The responsibility to maintain the model -- i.e. keep track of its accuracy and make sure it stays within the accepted threshold -- lies with the edge node that is also responsible to run it. The architecture is completely decentralized and the data center is there only to assist with the annotation. 

%\subsubsection{Federated Learning}
%In this paper, we do not consider \textit{Federating Learning}~\cite{mcmahan2017communicationefficient} to train the models. However, we have considered these techniques as part of a hierarchical edge architecture in which the automated annotation is carried out by models already tuned for nearby cameras. We believe the results from this work hint that our approach could benefit from federated learning in many aspects, but it remains out of the scope of this work.

%Moreover, in Section \ref{sec:baseline_model}, we analyze the impact of using as baseline a generic \textit{off-the-shelf} model -- i.e. trained on imagenet or other generic datasets -- with respect to a model that has already been trained on the same context of the deployment. 
%These results hint that our approach could largely benefit from federated learning techniques.

%\section{The COVA Method}
%\label{sec:problem_breakdown} 

\subsection{Considerations on Automatic Annotation}
The bigger models (potentially) have higher accuracy because they have more trainable parameters. When properly trained, the extra parameters can learn more and better features. These may come in the form of more layers, but the easiest way to increase accuracy (at the expense of computational cost) is to increase the input size~\cite{huang2017speed}. A larger input size increases the amount of information that the network receives in the form of pixels and allows models to \textit{see} objects bigger, meaning the network's receptive field takes more pixels into account which is particularly important when trying to detect small objects in the background.

\begin{figure*}[!ht]
    \centering
    \includegraphics[width=\linewidth]{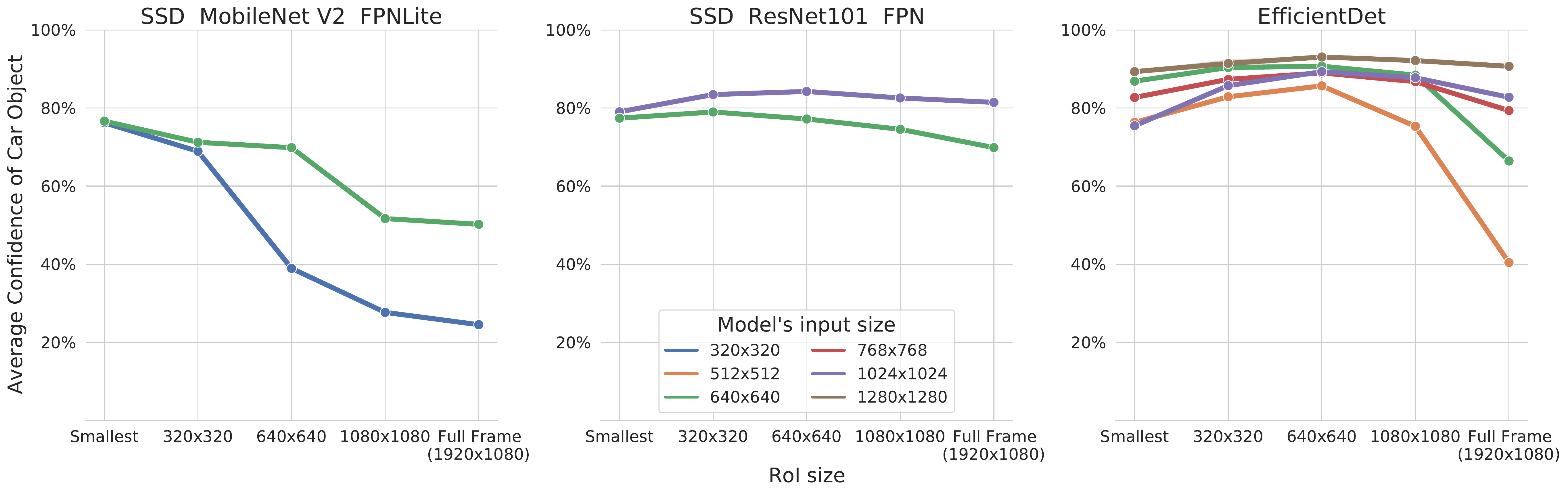}
    \caption{Average confidence of car detections from three models with different input sizes and varying the size of the Region of Interest (RoI) passed as inputs. All RoIs contain a car centered. \textit{Smallest} refers to the smallest square RoI that contains the car object, i.e., the smallest bounding box with aspect ratio of 1 that contains the full car object. \textit{Full Frame} refers to the full frame passed to the model as input at full resolution (1920x1080), i.e., as it was captured by the camera.}
    \label{fig:roi_size_accuracy}
\end{figure*}

Experiments show that mispredictions usually come from the model's inability to distinguish the object from the background (i.e., while Top-1 or Top-5 recall might be good, the confidence is below the threshold to be considered)~\cite{hsieh2018focus}. In other words, the model may correctly distinguish an object from other objects but not be sure whether the object is actually there. This phenomenon increases as the object gets smaller and becomes just a \textit{few pixels} large.
Figure ~\ref{fig:roi_size_accuracy} shows the impact on the confidence of predictions with respect to the size of the input image using different pre-trained models. The input images belong to different frames captured from one scene of the \textit{Racetrack} dataset, all containing a single car centered in the scene. For each frame, the region of interest (RoI) containing the car object is cropped in different sizes and fed to the model. Therefore, the car represents a smaller area of the model's input as we increase the RoI size. The experiment evaluates models with three different neural network architectures: SSD detector with MobileNet V2 feature extractor, SSD detector with ResNet101 feature extractor, and EfficientDet (also SSD with EfficientNet feature extractor). The models are different pre-trained versions of these three neural network architectures that were trained and dimensioned with different input sizes. 
%impact on the confidence of predictions with respect to the size of the input image, either enlarged or shrunk. %Results were obtained by passing a cropped RoI (Region of Interest) to the object detection model in which the object of interest appears on the RoI (Region of Interest).
These results show that confidence drops as the object gets smaller with respect to the rest of the scene. There are two factors contributing to this effect. First, the model's input size. Intuitively, bigger input sizes have a larger receptive field and are, potentially, better at locating small objects. Second, the neural network architecture. While the more modest MobileNet achieves on-par confidence compared to the state-of-the-art EfficientDet when the car represents the largest part of the input image, its confidence quickly drops as the object becomes smaller regardless of its input size. On the contrary, the ResNet101 and the EfficientNet backbones seem to sustain their confidence as the object gets smaller. However, their confidences still drop for the larger RoI sizes, especially for the models with smaller input sizes.
%\textcolor{red}{These results show how confidence varies as the object becomes bigger or smaller with respect of the rest of the image. \textit{No Minimum} refers to no enforcement of RoI size and the object will represent most of the image. From these results, we can extract that as the object becomes smaller, the model is less sure of it. However, it is interesting to see how without enforcing a minimum RoI size, the bigger model (ResNet) yields its worst results, probably due to the object being too zoomed in to the point it becomes bigger than what the network learned to identify during training.}

%\subsection{Determining the Right Size of the Training Dataset}
%How do we know we have enough images to train the model? There is no straight answer and all comes down to how much time we have allocated for this. A camera deployed into the wild to capture animals might have to wait weeks before one crosses its point of view while a camera counting cars in a busy intersection will have hundreds of examples in a matter of minutes. In Section \ref{tbd}, we analyze how the size of the training dataset impacts the resulting accuracy of the trained model.

%\subsection{Automatic Annotation of the Training Dataset}
%\sout{The edge node captures and sends images to the data center, where the \textit{ground-truth} model is running and waiting to run inference on those images. The predicted labels are sent back to the edge node, which will use them as ground truth during training.}

Finally, the \textit{ground-truth} model is assumed to be knowledgeable of the scope in which edge deployments will happen but is not required to be trained on images from those deployments. We build this assumption upon the observation that large models are better at retaining features that allow them to generalize better. These models are therefore expected to correctly identify \textit{most} of the objects from classes they have been trained to detect, even if these objects are seen from new perspectives and distances. This level of generalization is a characteristic that models small enough to run on edge devices do not share.  %We analyze this further in Section ~\ref{sec:generalization}.

%% file: framework.tex
\section{The COVA Framework}
\label{sec:framework}
COVA is a framework that automates the task of specializing neural network models for edge video analytics. As a framework, it aims to enable the rapid deployment of customized applications in resource-constrained edge nodes. As such, it has been implemented with customization and extensibility in mind.

COVA is made out of two main building blocks. On one side, it provides high-level structures to allow video analytics models to be tailored for a specific deployment in just a few lines of code. On the other side, it also provides multiple built-in methods to ease the extension and customization of the existing structures.

\subsection{Framework Architecture}
\label{sec:framework_architecture}

\begin{figure*}[!ht]
    \centering
    \includegraphics[width=\linewidth]{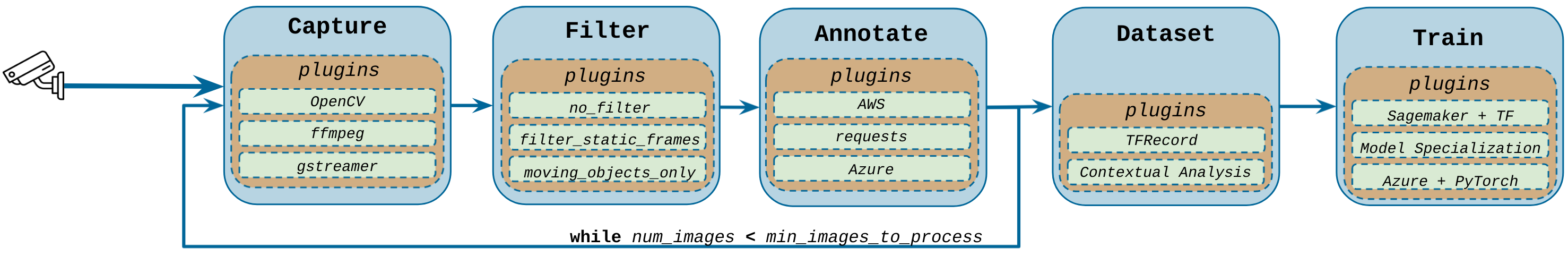}
    \caption{Default pipeline in COVA for automatic specialization of models.}
    \label{fig:autotune_workflow}
\end{figure*}

The backbone component in the workflow of COVA is the pipeline (or \textit{COVAPipeline}). A pipeline describes the sequence of steps involved in specializing a neural network model and how each step communicates with the next. Together, a sequence of steps conforms a pipeline. Figure~\ref{fig:autotune_workflow} depicts the default pipeline implemented in COVA entailing five steps. These are 1. capture images, 2. filter images, 3. annotate images, 4. create the dataset, and 5. training. The default pipeline iterates over steps 1-3 until enough images are available to continue. Then, step 4 generates the training dataset from the images and the annotations from the previous steps. Finally, the model is trained in step 5.

The potential of COVA relies on the high level of customization allowed by the pipeline architecture while keeping a simple structure within its core. Each step in the pipeline is implemented using a plugin architecture, allowing to easily extend or modify the default behavior. For example, the annotation step can be carried out using the built-in server exposing a REST API or the annotation plugin offloading the task to Amazon Web Services. Moreover, any step implementation (i.e., plugin) is interchangeable with any other implementation of the same step as long as their interfaces match what is described in the pipeline definition. Furthermore, it is possible to define different pipelines that use different stages and plugins and do so only through the configuration file, i.e., without any modification in the code.

\subsection{Pipeline Components}
By default, COVA implements a pipeline with the five stages previously described. However, the pipeline itself can be extended or modified if another sequence of stages is needed. Nonetheless, we describe and focus on the default pipeline, its stages, and the built-in plugins.

    \textbf{\textit{COVACapture}}. Captures and decodes images from a stream (input) and returns decoded frames as an RGB matrix. COVA implements capture using OpenCV, which, in turn, accepts different backends such as FFmpeg or GStreamer.
    
    \textbf{\textit{COVAFilter}}. Filters decoded frames (input) and returns a list of RGB matrices. Among the built-in filters, \textit{no\_filter} returns the same frame it received as input, i.e., does not apply any filtering; \textit{filter\_static\_frames} returns either the same input frame or an empty list, depending on whether any movement was detected in the latest frame or not; \textit{moving\_objects\_only} returns a list of the regions of the input frame where movement was detected.
    
    \textbf{\textit{COVAAnnotate}}. Annotates images. It receives filtered frames and returns the list of bounding boxes and labels of the objects detected on the input images. COVA implements two annotation methods: using the built-in REST API or using AWS. Both methods assume a ground-truth model trained on the target problem whose predictions are assumed to be ground-truths during training. Section~\ref{sec:results} evaluates the impact of this assumption. This stage supports both TensorFlow and OpenVINO models.
    
    \textbf{\textit{COVADataset}}. Generates the training dataset from the filtered images and their respective annotations. The built-in plugin results in a TFRecord file as output (format used in TensorFlow that stores data as a sequence of binary strings). However, the resulting dataset can be subject to different processes. Specifically, COVA implements a default dataset in which all annotations make it into the dataset (since filtering occurs in the Filter stage). %Additionally, COVA provides another method called \textit{Contextual Analysis} that can be used to generate highly-specialized binary classifiers whose labels are decided upon a previous analysis of the contextual features present in the scene. This method is introduced in Section~\ref{sec:discussion}.
    
    \textbf{\textit{COVATrain}}. Trains a neural network using the previously generated training dataset. TensorFlow is the default library used. Built-in plugins support two flavors of training: standard TensorFlow and TensorFlow Object Detection API (used for training or fine-tuning object detection models). Both flavors can be executed either locally or using AWS. As output, this stage generates a specialized model (in TensorFlow's \textit{SavedModel} format).

\subsection{Pipeline Configuration}
COVA allows pipelines to be fully defined using a configuration file in JSON format. The configuration file allows users to dynamically define the plugins to be used on each pipeline stage and the parameters expected by each stage. Figure~\ref{lst:config_file} shows an example of a pipeline defined using a configuration file in JSON format. Custom plugins can be loaded in the configuration file using the \textit{plugin\_path} keyword. The set of parameters accepted by each plugin will ultimately depend on the specific plugin implementation. However, some of the most important ones include:

\begin{figure}[ht]
    \includegraphics[width=0.7\linewidth]{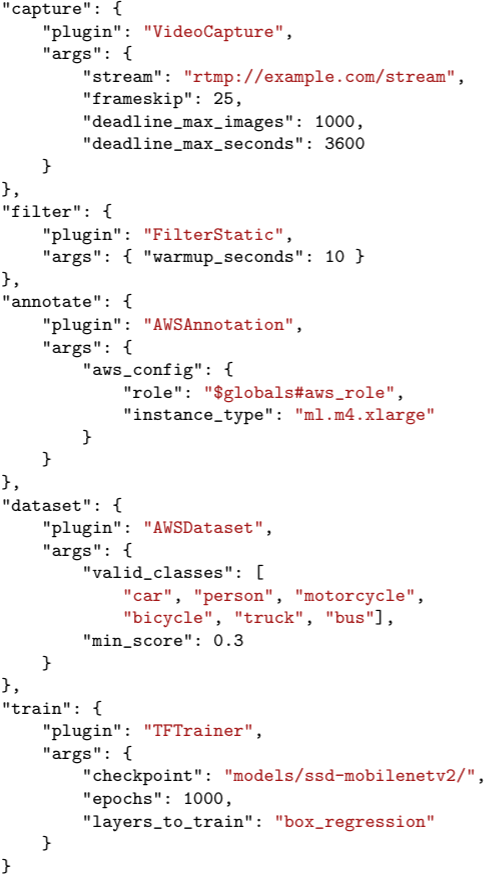}
    \caption{Example of configuration file defining a COVA pipeline.}
\label{lst:config_file}
\end{figure}

% COVA assumes three inputs. First, one or many video streams whose images will be used to train models. Second, the Edge neural network to train. It can be uninitialized or pre-trained. If pre-trained, its weights will be used as baseline and the model will be fine-tuned. Third, COVA assumes access to a ground-truth model that will be used as oracle during the annotation of the training images. It is assumed that the ground-truth model is already trained on a superset of the problem at hand.
% The framework assumes an initial state with the following components:
% \begin{itemize}
%     \item Video feed whose images will be used to fine-tune models.
%     \item Edge model to use as baseline. This is the model that will be fine-tuned.
%     \item The ground truth model that is used as oracle to annotate.
% \end{itemize}

    \textbf{Target classes of objects}. List of object classes for which the model will be specialized. The list of classes is the basis of model specialization for two reasons. On the one hand, the type of objects expected to be seen and detected in the scene is essential to define a camera's context. Limiting the classes of objects the model has to learn to identify helps specialize the resulting model, and the learning process can focus on those classes it will eventually encounter. On the other hand, the precision of the ground-truth model gets a boost after we discard detections from classes that are unlikely to be seen in a specific context (e.g., a boat detected crossing a pedestrian crossing can be omitted or, at the very least, flagged to be double-checked). Only frames containing instances of one of the target classes are considered during the dataset's creation, while the rest are filtered out.

    \textbf{Training deadline}. It is important to set a deadline to force the trigger of the training process in scenarios where there is no high influx of new objects of interest entering the scene. If the training is triggered by the deadline, results may not be optimal, but, at least, the model can start working while new images are being captured for a new training iteration.
    
    \textbf{Motion sensitiveness}. If the filter plugin uses motion detection, it can be tweaked to be more or less sensitive to small changes in the frame. This parameter will highly depend on the characteristics of the scene. The smaller the objects we want to capture, the higher the sensitivity required to detect their movement. However, high sensitivity may incur an undesirably large number of frames being flagged with motion, as small variations can occur due to glitches, camera jitter, or the wind blowing on background objects.
    
    \textbf{Trainable Layers}. If the training plugin uses a checkpoint of a pre-trained model, COVA allows defining the layers that are to be trained while the rest are frozen. We have tested three configurations: \textit{unfrozen} (all weights are trainable), box regression (only the layers of the object detector head are trainable), and top (box regression plus the last few layers of the feature extractor). According to our experiments, the optimal configuration will highly depend on the size of the training dataset. With enough examples, unfrozen seems to yield the best results. However, box regression is the safest option and yields good results even with only tens of images, while unfrozen has a higher risk of overfitting with small datasets.

\subsection{Built-in Utilities}
COVA also provides several built-in methods aimed at assisting with the development of new applications. Among all the utilities and auxiliary methods, we primarily focus on motion detection, as it is the basis of how COVA analyzes the scene to provide better results while reducing the amount of network bandwidth (as only a portion of the frame needs to be sent) and compute used (by reducing the amount of annotations required).

\subsubsection{Motion Detection}
% \textcolor{red}{MOTION DETECTION OR BACKGROUND SUBTRACTION?}
There is a myriad of existing methods to perform motion detection. Nonetheless, it involves many challenges (e.g., camera jitter or lighting variations), and, as a problem, it is far from being solved. There is not a \textit{de-facto} approach that can handle all scenarios robustly, and different methods offer different trade-offs in terms of speed or accuracy. In COVA, we prioritized speed over accuracy when selecting the methods to implement for two reasons. First, the accuracy of these methods is usually evaluated using image quality metrics of the generated background or error metrics at the pixel level to compare the generated background to the ground-truth background. Second, COVA must run on resource-constrained Edge nodes and do it faster and more cost-effectively than the alternative: send whole frames over the network and process them in a centralized data center. Therefore, COVA implements two simple methods consisting of two steps or building blocks:
%For example, OpenCV implements out-of-the-box several methods for background subtraction~\cite{opencv_background_subtraction}. However, these methods have a relatively high computational cost as they aim to accurately identify each foreground object, which is similar but not the same problem we want to solve. Our goal is to detect regions of the frame that changed with respect to the background but whatever is not moving can be safely assumed to be background, as we assume it was detected the moment it entered the scene. Therefore, COVA implements a simple method consisting of two building blocks:

\textbf{Background Modeling of the scene} is the key step for motion detection to work effectively. Take an incorrect background of the scene, and any advantage from motion detection fades away, as every single frame could be incorrectly flagged as containing changes. As previously mentioned, COVA assumes images from static cameras, which greatly simplifies the background modeling.
Nonetheless, the optimal algorithm will still be highly dependant on the specific context in which the camera is placed. Therefore, COVA implements several methods. The first and simplest is to assume that the first frame captured is the background of the scene. This method works best when the scene is known to be empty of moving objects when the process starts, and the background is not subject to long-term changes. However, in many cases, this is not only false, but the definition of background can change over time (e.g., sunlight turning into darkness or objects left unattended for long periods, like parked cars). Therefore, background modeling works best when the background is computed over time. In this direction, COVA implements background modeling using a Mixture of Gaussians (MoGs)~\cite{stauffer1999adaptive}. Our results have shown that this method yields robust results while being computationally inexpensive (i.e., on average, 2.7 milliseconds on a single core of an Intel Xeon 4114). This is the method used by default.
% implements two methodologies: the pixel by pixel moving average of the last \textit{N} frames (considering some frame skipping). We previously considered using average values but we detected that the median -- or percentiles -- yields more robust results, as objects that stop within the scene for enough frames created artifacts that stayed for long periods of time until their effect was removed from the average. Using median values will integrate stationary objects to the background -- which is desirable -- but remove them again short after they leave the scene.

\textbf{Monitor the scene for any substantial changes}. A simple subtraction operation will give us the pixel-to-pixel absolute difference between the current frame and the background model. Nonetheless, in real-world deployments, two consecutive frames will never be identical. Multiple factors will cause small pixel changes even with a static background and no objects moving. 

On the one hand, digital camera sensors are imperfect and introduce tiny variations that cause pixel values to vary, even when the scene remains seemingly invariant. Applying a Gaussian blur to both the background and the current frame filters noise out when computing the pixel-to-pixel difference. Next, COVA applies a binary threshold to the computed difference to consider only those regions containing significant changes in the pixel intensity. After applying a \textit{dilate} operation to fill in the holes over the result of the binary threshold, we obtain a mask with the changing regions in the frame. Finally, we compute the bounding boxes by finding the contours on the masked frame. A contour is a curve joining all the adjacent points along the boundary of a region of pixels with the same color or intensity.

On the other hand, small changes can still be undesirably (although not necessarily incorrectly) flagged as motion. There is no rule of thumb to determine how large a region must be to be considered a potential moving object, as it will highly depend on the specifics of the scene and the objects we want to detect. For example, in the \textit{Racetrack} dataset, some cameras point directly to the track at a relatively short distance, and cars are seen big enough to consider only large regions as motion. However, other cameras have a broader view and capture cars and other objects from different distances, and, area-wise, a distant car might be indistinguishable from a leaf fluttered by the breeze from a nearby tree. Therefore, COVA considers a user-defined threshold that sets the minimum area for regions to be considered of interest.

%COVA applies the following operations to compute regions of interest: first, it applies Gaussian blur to the frame to filter noise out. Then, it obtains the absolute difference between frame and background (also blurred) and applies a binary threshold to consider only regions with significant changes in the pixel intensity values (pixels with values below and above the threshold are set to black and white, respectively). Finally, it applies a \textit{dilate} operation to fill in the holes after the binary threshold.

% On the other hand, shadows cast by background objects move over time, or the wind blowing may shake the camera itself. To avoid these situations to invalidate motion detection, COVA can be configured with a timer to reset the background. This is specially important during the annotation phase, as each frame incorrectly misplaced as containing changes will have cost associated. 

\begin{figure} 
    \centering
    \subfloat[]{%
        \label{subfig:track_bbox}
        \includegraphics[width=0.45\linewidth]{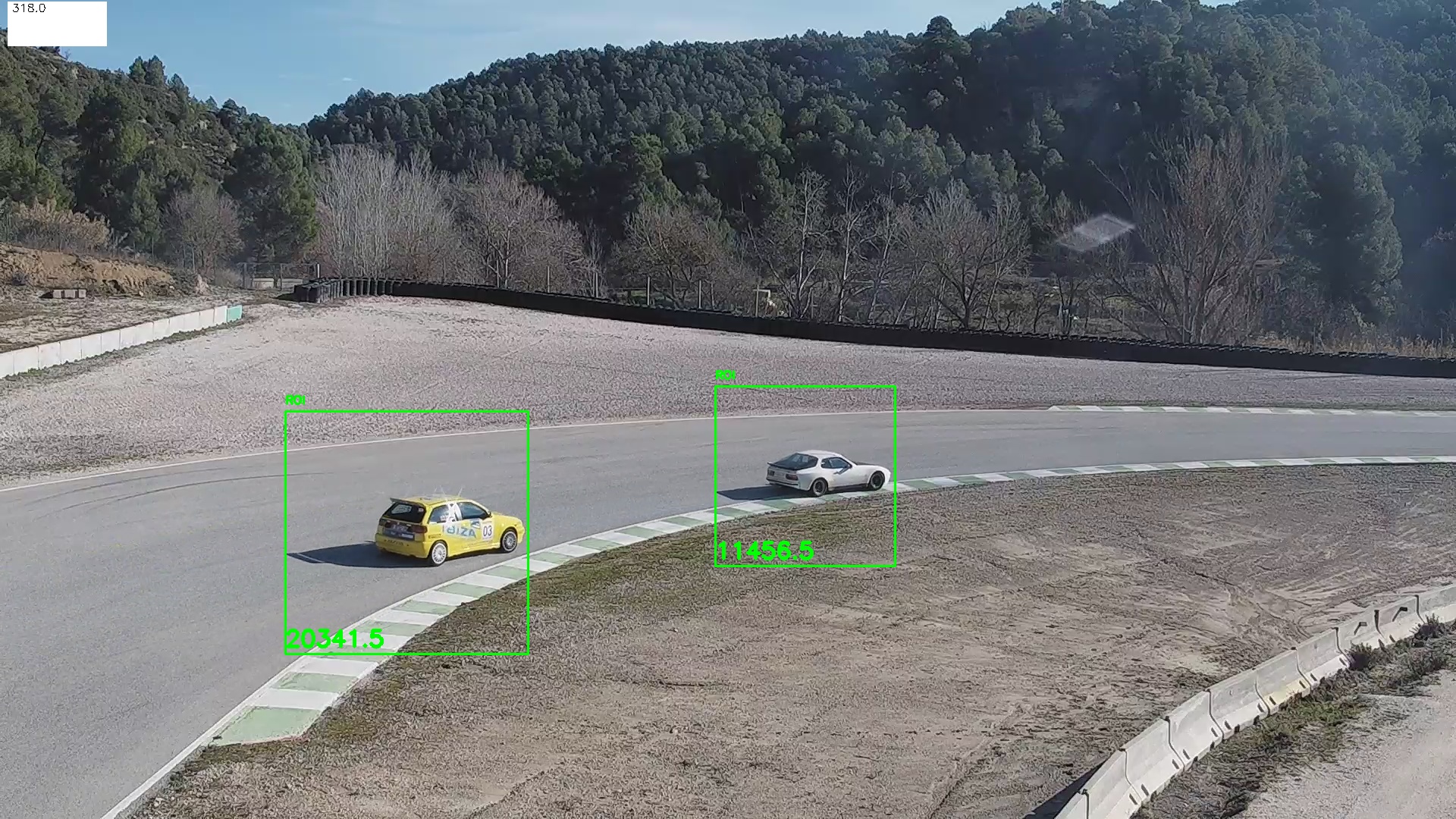}}
    \hfill
    \subfloat[]{%
        \label{subfig:tack_threshold}
        \includegraphics[width=0.45\linewidth]{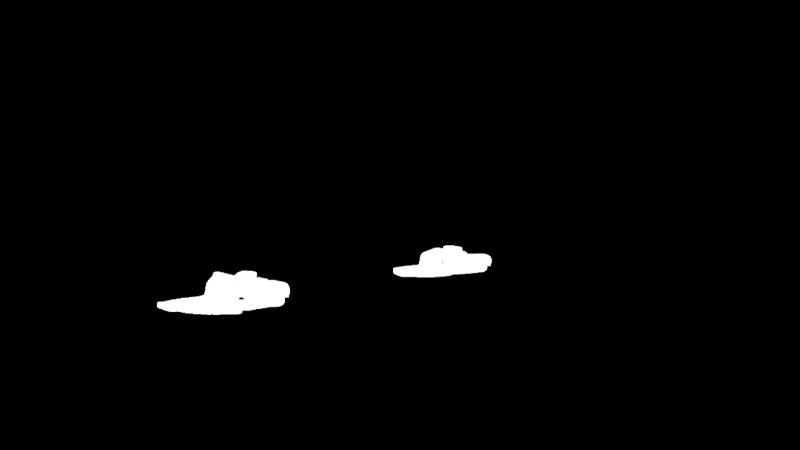}}
  \caption{(a) Two cars in the racetrack with bounding boxes around them, obtained after performing background subtraction on the frame. (b) Thanks to a static background, the task of separating moving objects from the background is largely simplified. The regions of interest after applying simple computer vision techniques are much smaller than the whole FullHD frame captured by the camera lens, which allows us to focus the attention closer to the objects and increase the accuracy of the ground-truth model. In an scenario like the racetrack of the image, motion detection alone is able to filter all empty frames out and only consider those with cars and minimize the annotation costs.}
  \label{fig:track_motion} 
\end{figure}

Figure \ref{fig:track_motion} depicts an example of the process, where two cars are correctly detected as the only two objects moving in the scene.
Each one of these operations is already implemented by OpenCV~\cite{opencv_background_subtraction}. 
Altogether, they result in an inexpensive yet effective method for our purpose. Through them, motion detection takes orders of magnitude less time than computing a single inference on resource-constrained nodes. Therefore, it becomes ideal to be executed on resource-constrained nodes on a frame-by-frame basis to optimize the automatic annotation of images, which is indeed the most expensive single operation in the whole process.

%It can be noted that the modeling of the background is the key step for this method to effectively work. With an incorrect background of the scene, the method becomes useless, as every single frame could be incorrectly marked as containing changes. Therefore, we must first obtain the background of the scene. As previously discussed, static cameras have the peculiarity that the background is \textit{mostly} static and does not change over time, at least at a small time scale. We obtain the background by taking the pixel by pixel median of \textit{N} frames. Then, we only have to do a simple subtract operation between the current frame and the background to obtain the differences. However, a simple difference between consecutive frames may yield small changes that could jeopardize the effectiveness of the method. 

\subsubsection{Cloud-Assisted Edge Deployments}
As part of the Edge-Cloud interplay that COVA aims to leverage, some of the built-in plugins already implement the usage of Amazon Web Services. Specifically, S3 to upload images and \textit{SageMaker} for the annotation and training stages. Moreover, COVA provides several auxiliary modules and utilities to integrate these services on new applications.

\subsection{Source Code Availability}
The entirety of the COVA project has been made publicly available in the project's repository~\cite{edge_repo} with an open license (Apache License 2.0) to encourage the community to use it, test it, and extend it. The repository includes the fully documented source code with detailed instructions on deploying it end-to-end and reproducing the experiments described in this paper.

COVA is implemented in Python. By default, its only requirements are OpenCV and TensorFlow, while CUDA can be used if available.

% \begin{itemize}
%     \item OpenCV. Video processing and manipulation of images.
%     \item TensorFlow 2.4. For anything related to training or execution of neural networks. PyTorch is also supported for the execution of neural networks during annotation by the ground-truth model.  
%     \item Flask. RESTful API to offload annotation to the data center.
% \end{itemize}

%% file: methodology.tex
\section{Methodology}
\label{sec:methodology}
In this section, we detail the methodology followed throughout the experiments to evaluate their results. 

\subsection{Edge Model Taxonomy}
The evaluation in Section~\ref{sec:results} is carried out using two neural network architectures and three different versions of each one. Table~\ref{tab:model_taxonomy} describes the model taxonomy used throughout the remainder of the paper. The \textit{edge model} refers to the model that we aim to optimize (i.e., specialize) for its deployment to the edge. It is typically a lightweight model that can run at real-time performance on resource-constrained nodes. The \textit{ground-truth model} refers to the model whose predictions are considered to be ground-truths while building the dataset to use during the training phase of the edge model. Edge and ground-truth are conceptual classifications for two models that play different roles within our proposed solution. Therefore, these do not refer to any specific neural network architectures, and any model can be used as either edge or ground-truth models. However, the ground-truth model is expected to be more accurate and more complex than the edge model. Otherwise, the latter would be a better choice to deploy to the edge. 

\begin{table}[]
\begin{tabular}{rcl}
\multicolumn{1}{c}{\textbf{Model}}                & \textbf{Variant} & \multicolumn{1}{c}{\textbf{Trained/Fine-Tuned on}} \\ \hline
\multicolumn{1}{r|}{\multirow{2}{*}{ground-truth}} & off-the-shelf    & pre-trained on MS COCO                             \\
\multicolumn{1}{r|}{}                             & generic          & other similar edge scenarios                                 \\
\multicolumn{1}{r|}{\multirow{3}{*}{Edge}}        & off-the-shelf    & pre-trained on MS COCO                             \\
\multicolumn{1}{r|}{}                             & generic          & other similar edge scenarios                                 \\
\multicolumn{1}{r|}{}                             & specialized       & the same camera of deployment                                   
\end{tabular}
\caption{Summary table of the model taxonomy with the different types of models considered throughout the paper.}
\label{tab:model_taxonomy}
\end{table}

Moreover, we define three different variants that refer to how the model was trained: \textit{off-the-shelf}, generic, and specialized. Off-the-shelf (OTS) refers to the model pre-trained on the COCO dataset (Common Objects in Context)~\cite{lin2014microsoft} without any further training nor tuning on any other context. It is easy to find most of the well-known neural network architectures already trained on this dataset.
Generic and specialist variants are categories always relative to the model to deploy. Generic refers to a version of the model trained or fine-tuned on contexts similar to the deployment context. For example, if we want to deploy a model to control cars at a parking lot, a generic model could have been trained using images from other parking lots or urban areas with mostly cars. Specialized refers to a model that has been trained using images from the same edge camera of the deployment.

\subsection{Datasets}
\label{sec:datasets}
Throughout this paper, we use two datasets to evaluate the different points of our approach. 

\begin{itemize}
    \item VIRAT Video Dataset~\cite{oh2011large}. Data collected in "\textit{in natural scenes showing people performing normal actions in standard contexts, with uncontrolled, cluttered backgrounds}". After filtering scenes that were not fully annotated, the dataset we used contains a total of 10 scenes from 10 different static cameras. We have grouped these 10 scenes into 4 contexts: \textit{Campus Inside} for scenes 0000, 0001, and 0002; \textit{Campus Outside} for scenes 0100, 0101, and 0102; \textit{Parking Lot} for scenes 0401 and 0403; and \textit{Street} for scenes 0501 and 0503. The name of these groups is descriptive of the context in which the images were captured.
    \item Racetrack. Custom dataset with footage captured during car training session in a racetrack. The dataset includes images from 6 static cameras placed in different spots along the track and captured at 12 different points in time. % from 9:30h to 17:00h. 
    This dataset is not fully annotated, and we use it only for a meta-analysis of its characteristics. It is nonetheless interesting, as it gives the perspective of a use case in which cars are virtually the only moving objects crossing the field of vision of the cameras.
\end{itemize}

In the experiments presented in Section~\ref{sec:results}, we focus the evaluation on the VIRAT dataset, as we consider it to present enough diversity of scenes to be representative.

\subsection{Evaluation Metrics}
In image classification, the two main metrics to evaluate a model are \textit{precision} and \textit{recall}. For a certain class of objects, precision will tell us how much we can trust that a given prediction has been correctly classified (i.e., the ratio of True Positives with respect to the total of predictions), while recall will tell us how much we can trust that images of a certain class are correctly classified (i.e., the ratio of True Positives with respect to the total number of \textit{ground-truth} True Positives). However, the goal of object detection models is to find the bounding box coordinates of each detected object, and these coordinates are real numbers. Therefore, to classify a prediction as True or False Positive and Negatives, we consider the Intersection over Union (IoU) between the predicted box and the ground-truth. The detection is considered a True Positive if the computed IoU is over a certain threshold and is considered False Positive otherwise.

\subsubsection{Mean Average Precision (mAP)}
We use mAP (mean Average Precision) as the main metric to evaluate each version of the different models resulting from the experiments. Within object detection, mAP has gained popularity as it is used in some of the most popular challenges in the field, such as PASCAL Visual Object Classes (VOC)~\cite{everingham2010pascal} and MS Common Objects in Context (COCO)~\cite{lin2014microsoft}. However, the details of how mAP is computed may vary from challenge to challenge. In this paper, we compute mAP as specified by the COCO challenge.

\subsection{Experiment Setup}
The experiments explore the different considerations we have identified. However, there are a series of parameters that are common throughout the different experiments.
\begin{itemize}
    \item Transfer learning affects only the parameters on the box regression layers of the model, i.e., box regression layers are left unfrozen, while feature extraction layers are frozen.
    \item Size of the dataset: collection stops at 1000 training images \textit{per scene}. The evaluation dataset contains two hundred images per scene. Training and evaluation images are extracted from different videos, i.e., they contain images from different points in time to reduce overlapping.
    \item Models are trained for 5000 epochs. 
\end{itemize}

\subsubsection{Baseline OTS Models}
For the evaluation, we start from an edge and ground-truth \textit{off-the-shelf} models. These models are publicly accessible from the TF Model Zoo~\cite{repo:tf_modelzoo} and have been pre-trained on the COCO (Common Objects in COntext) dataset~\cite{lin2014microsoft}. For the experiments, we have selected a MobileNetV2 feature extractor with a Single Shot Detector (SSD) detection head~\cite{repo:edge_model} with an input size of 300x300x3 pixels as the \textit{edge} model and an EfficientNet feature extractor with an SSD head~\cite{repo:ref_model} and an input size of 768x768x3 pixels as our \textit{ground-truth} model. The edge model was selected for being the smallest and fastest pre-trained model available with a  while the ground-truth model was selected for being on the high-end of best performing models. The mAP of the models on the COCO dataset is 20.2\% and 41.8\%, for the edge and ground-truth models, respectively.

%% file: results.tex
\section{Results}
\label{sec:results}

We have broken down the process of automatic tuning of neural networks into different steps. Each step invariably brings a trade-off that we must understand before moving forward.
% Before transitioning towards an unsupervised system to fine-tune and adapt neural networks to their deployment contexts, we must understand the trade-offs at each step. 

The following experiments are designed to test the hypotheses on which COVA is based. First, we test the level of generalization \textit{off-the-shelf} achieved by the ground-truth model compared with the edge model. Then, we analyze the potential impact of specialization on accuracy by comparing the accuracy of generalist models with that of models specialized for a specific scene or context. Finally, we evaluate the impact that an automatic annotation of the training dataset has on the accuracy of the resulting specialized models, i.e., the accuracy drop with respect to manual annotation, which is assumed to be perfect. 
%The section is divided as follows:

% \begin{enumerate}
%     \item Off-the-shelf generalization. First, we start by testing how our ground-truth and edge models generalize off-the-shelf, i.e. pre-trained on generic datasets such as ImageNet and MSCOCO, which include a superset of classes that we will be testing on our specific datasets
%     \item Generic models vs Specialized models. In this experiment we evaluate the level of generalization that the edge model can achieve after being trained on similar contexts to the context to which it will be deployed and this is compared against a specialized model that has been trained on a subset of images from the same context on which it is evaluated.
%     \item Automatic annotation. In this experiment, we evaluate the results obtained by an edge model that is trained on images annotated by a ground-truth model. 
% \end{enumerate}

\subsection{Off-the-Shelf Generalization}
This experiment tests the hypothesis that larger models are able to generalize better than smaller models, i.e., achieve higher accuracy without being fine-tuned or trained on the specific context on which we evaluate them.

\begin{figure}[!ht]
    \centering
    \includegraphics[width=\linewidth]{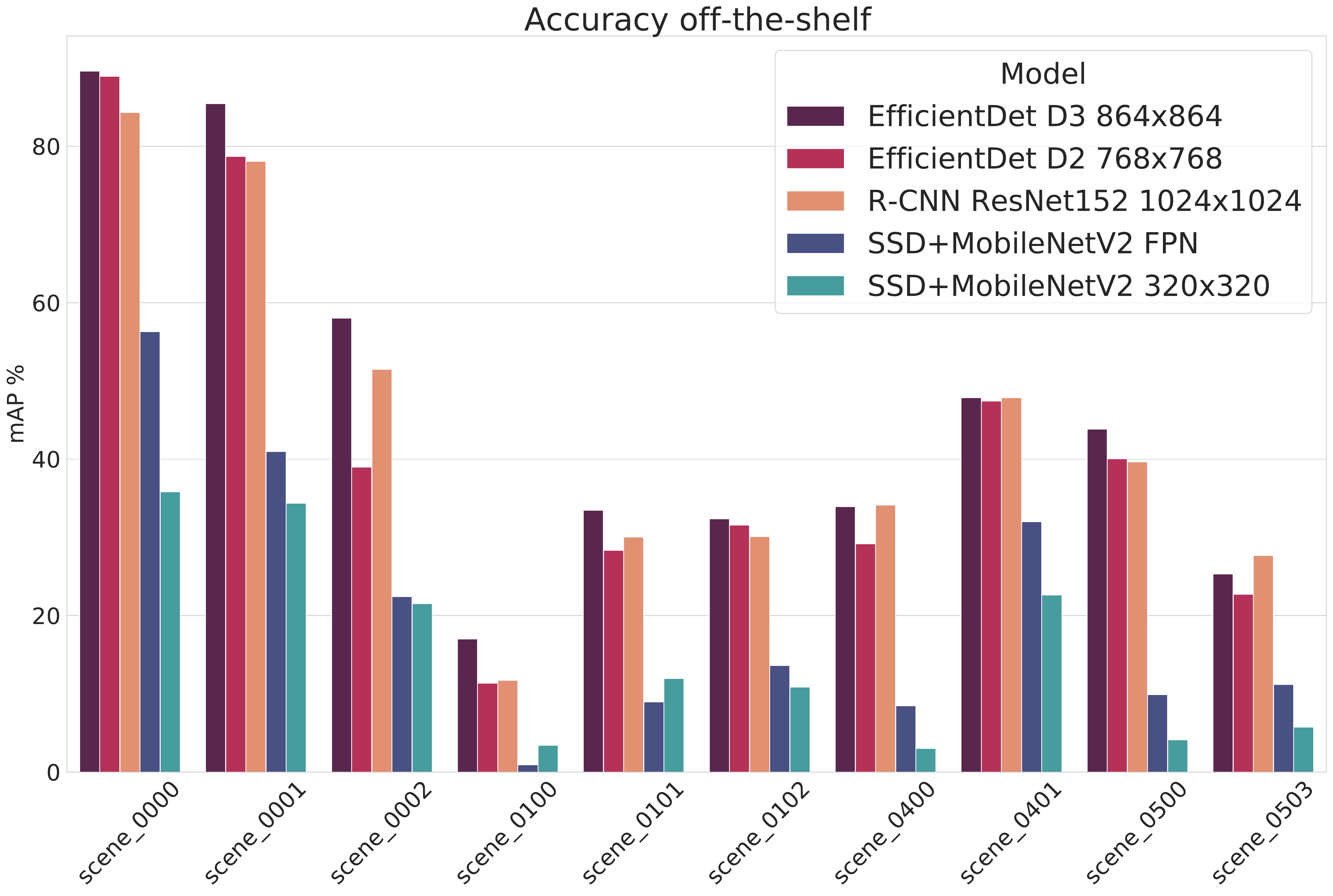}
    \caption{Mean Average Precision of three ground-truth models and two edge models \textit{off-the-shelf} evaluated on the VIRAT dataset. Breakdown by scene.}
    \label{fig:base_model}
\end{figure}

Table~\ref{tab:map_ots} shows the average mAP achieved by different ground-truth and edge models \textit{off-the-shelf}, i.e., pre-trained on the COCO dataset. All three ground-truth models evaluated achieve similar or higher scores than what they achieved on the dataset on which they were trained. However, both edge models struggle to generalize to the VIRAT dataset using their pre-trained knowledge and drop their accuracy between 8\% and 25\%. Figure ~\ref{fig:base_model} zooms into the mAP of all models broken down by scene. Again, all ground-truth models consistently outperform the edge models on all scenes. The difference between ground-truth and edge models fluctuates between 2x and 18x, depending on the scene. On average, the ground-truth models outperform the edge models by a factor of 6x. These results highlight two things. First, the ground-truth models achieve significantly better accuracy than the smaller edge models, as expected. Second, ground-truth models seem to sustain their accuracy in new environments, implying a higher generalization level.

\begin{table*}[t]
\centering
\begin{tabular}{rcccc}
\textbf{Model}          & \textbf{Category} & \textbf{COCO mAP} & \textbf{VIRAT mAP} & \textbf{Diff} \\ \hline
EfficientDetD3          & Ground-truth      & 45.4              & 46.7               & +3\%          \\
EfficientDetD2          & Ground-truth      & 41.8              & 41.7               & -0.25\%       \\
R-CNN ResNet152         & Ground-truth      & 39.6              & 43.5               & +10\%         \\
SSD MobilenetV2 FPNLite & Edge              & 22.2              & 20.5               & -8\%          \\
SSD MobilenetV2         & Edge              & 20.2              & 15.4               & -25\%        
\end{tabular}
\caption{Comparison of mean Average Precision on the COCO and VIRAT (average of all scenes) datasets for different \textit{off-the-shelf} ground-truth and edge models.}
\label{tab:map_ots}
\end{table*}

\subsection{Generic vs Specialized}
This experiment tests the hypothesis that a specialized model performs better than a generic model when evaluated on the same context of its deployment.

We define the \textit{generic} edge model as a version of the edge model that has been fine-tuned using images from scenes similar to the one we want to deploy the model. That is, images from other cameras different than the one to evaluate but belonging to the same supercategory as used in Section~\ref{sec:problem_breakdown} (i.e., Campus Outside, Campus Inside, Parking Lot, Street). Analogously, we define the \textit{specialized} edge model as a version of the edge model that has been fine-tuned using images from the context to deploy. That is, the model is fine-tuned using images from a single camera. Nonetheless, images on the training and evaluation datasets are from videos captured at different times, i.e., a video is used for either training or evaluation dataset, but never in both.

\begin{figure}[!ht]
    \centering
    \includegraphics[width=\linewidth]{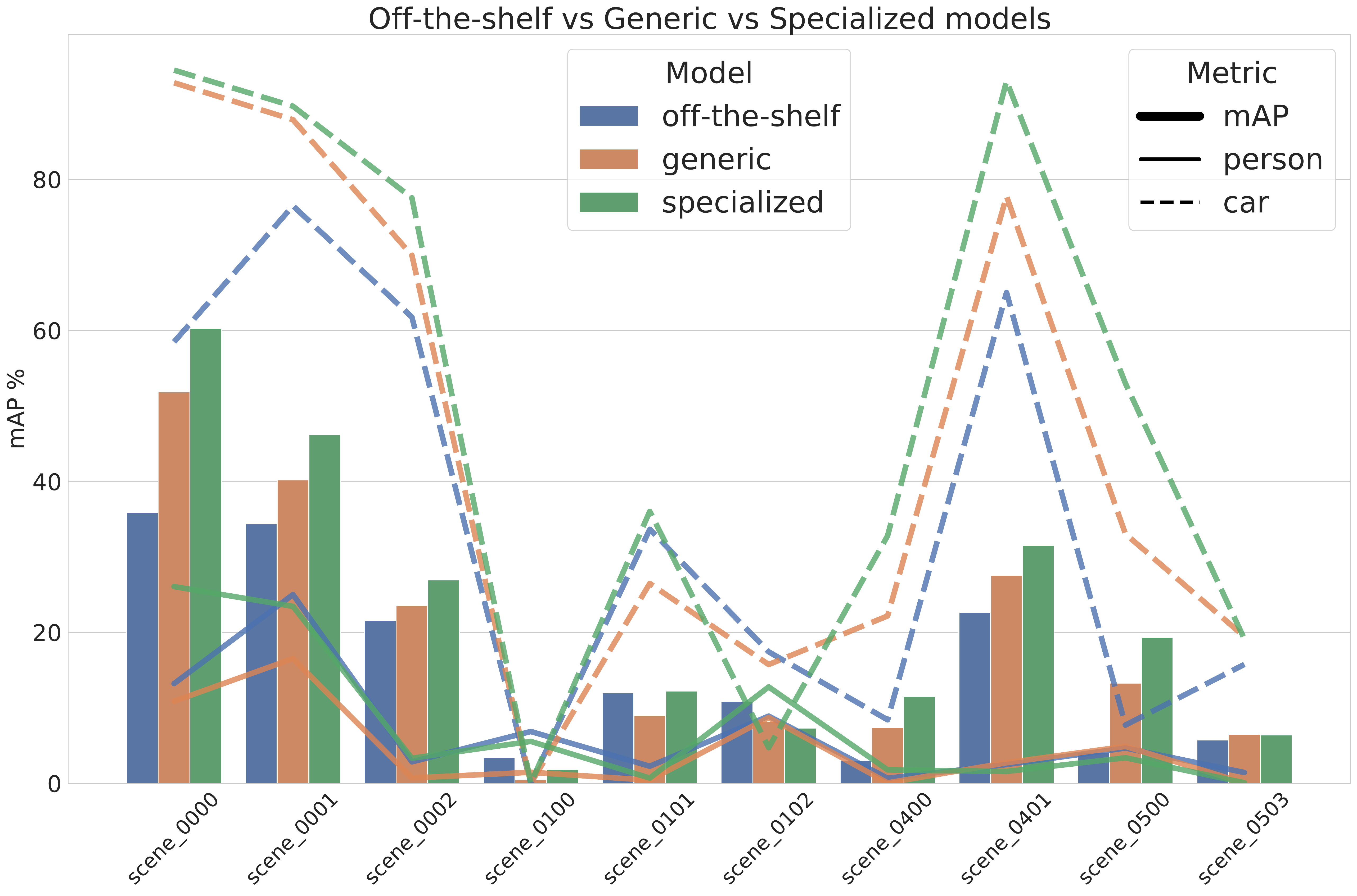}
    \caption{Mean Average Precision of the generic edge model compared to the specialized edge model for the VIRAT dataset. Breakdown by scene. Bars show average mAP and lines and dotted lines show person and car AP, respectively.}
    \label{fig:generic_vs_special}
\end{figure}

Figure ~\ref{fig:generic_vs_special} shows the mAP of the generic and the specialized edge models. On the one hand, the \textit{off-the-shelf} models consistently underperform both generic and specialized models except on one scene. In scene 0102 the pre-trained model achieves 10\% higher mAP than the generic and 15\% higher than the specialized. Diving into the breakdown accuracy by object classes, the specialized achieves higher accuracy on person detection but still drops the average accuracy due to poor car detection. Upon closer inspection of the contents of the scene, we observed that the scene (supercategory \textit{Campus Inside}) contains a single instance of a car, which was seemingly insufficient during training.
On the other hand, the specialized model outperforms the generic model on every scene of the dataset (except, again, scene 0102) by a minimum of 2\% and up to 87\%, with an average improvement of 30\%. These results prove how model specialization can help us achieve greater accuracy without altering the computational complexity of the model. However, they also highlight that special care should be taken in scenes where there is an imbalance on the class distribution of the objects captured.

\subsection{Manual Ground-truth vs. Automatic Annotation}
This experiment evaluates the impact of automatic annotation on the resulting model's accuracy. The edge model is trained using the same set of examples whose annotations were obtained using three annotation variants. On the one hand, manual annotation is taken as the reference, as it is assumed to be perfect. On the other hand, automatic annotation in which predictions from the ground-truth model are taken as ground-truths during training. Automatic annotation is evaluated with a minimum confidence threshold of 0.3 and 0.5. Therefore, predictions from the ground-truth model with lower confidence are discarded.
Moreover, the ground-truth model is assumed to be trained on (or, at least, knowledgeable of) the classes of objects that will have to annotate and detect. %We have annotated the VIRAT dataset using the ground-truth model and accepted any detected object with confidence above a given threshold as ground-truth. 

\begin{figure}[!ht]
    \centering
    \includegraphics[width=\linewidth]{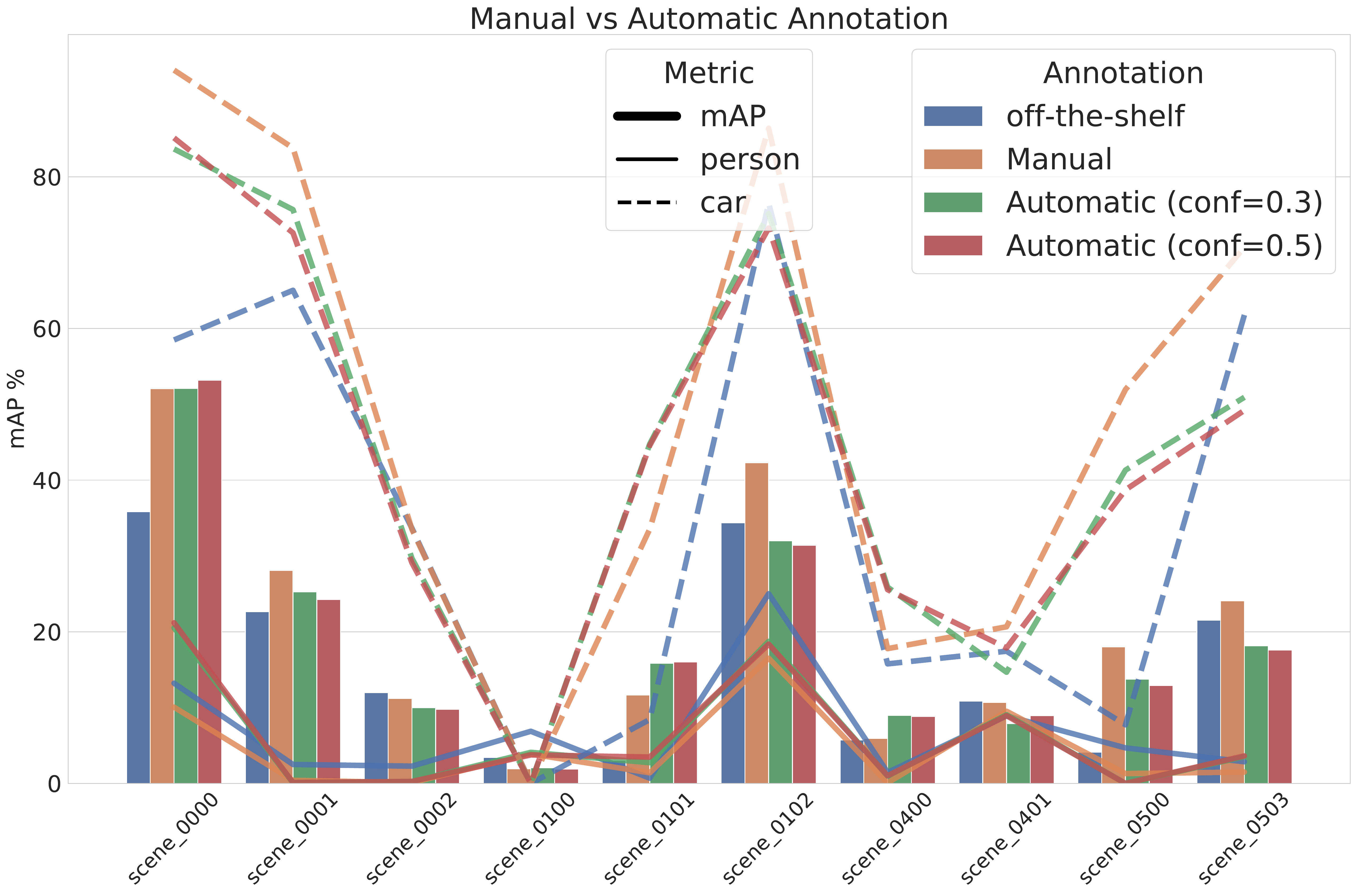}
    \caption{Mean Average Precision of a specialized edge model after being trained on a manually annotated dataset vs. a dataset automatically annotated by the ground-truth model with thresholds of confidence of 0.3 and 0.5. Bars show average mAP, and lines and dotted lines show person and car AP, respectively.}
    \label{fig:manual_vs_special}
\end{figure}

Figure \ref{fig:manual_vs_special} shows the mean Average Precision of the specialized edge model using three datasets annotated differently: manual, automatic (conf=0.3) and automatic (conf=0.5). Compared to manual annotation, the automatic annotation yields similar results on some scenes while it lags behind on others. Upon closer inspection of the different metrics, car objects' weight in the scene seems to be directly related to the quality (i.e., mAP) of the resulting models trained using automatic annotation. A closer look at the composition of the scenes where automatic annotation lacks significantly behind manual annotation hinted that the problem lies in the smaller objects that dominate those scenes, like person instances. Small objects are usually more difficult to detect for obvious reasons, and this translates to second-order errors, as the ground-truth model introduces annotation errors (mostly in the form of False Negatives by not detecting them). Consequently, the edge model fails to detect them during inference due to size and deficient training. However, automatic annotation successfully outperforms the results of the off-the-shelf model and increases the mAP of all scenes in average by 21\%. It is worth noting that varying the confidence threshold from 0.3 to 0.5 does not result in substantial differences. This seems to be due to the low number of False Positives introduced during the annotation of the dataset.

% Consequently, for the following experiments we used a fine-tuned version of the ground-truth model, equivalent to the generic edge model -- i.e. trained on all scenes but the one we evaluate. 

%Therefore, we can establish that the previous results for the generic edge model set the lower bound accuracy of our solution, i.e. these are the numbers to beat, since we need the same set of images to train the generic edge and generic ground-truth models. If a specialized model trained on images annotated by a generic ground-truth model is not capable of outperforming a generic model, then the automatic fine-tuning of specialized models has failed on its goal and training would be better spent on generic edge models than on generic ground-truth models that will later require multiple inferences to annotate other datasets. Consequently, the previous accuracy results for the specialized edge model trained on grountruth annotations set the upper bound for our solution.

%% file: conclusions.tex
\section{Conclusions and Future Work}
\label{sec:conclusions}
In this paper, we have presented and evaluated COVA (Contextually Optimized Video Analytics), a framework that automates the process of model specialization to ease the deployments of real-time video analytics on resource-constrained edge nodes. COVA provides various high-level structures and tools that allow users to quickly define and customize the automated pipeline, resulting in object detection models specially tailored for their deployments.

COVA results from an extensive exploration and a subsequent analysis of the considerations to optimize the execution of models in resource-constrained edge nodes. We have successfully identified several key characteristics of edge video analytics that allowed us to efficiently and largely simplify the scope of the problem. For example, our results have shown that simple motion detection techniques can boost the accuracy of already trained models by allowing us to direct the inference to those parts of the frame that are moving. 

Moreover, COVA allowed us to explore some of these considerations, which we hope can help pave the way for future improvements. We have shown how, in the case of static cameras, it is worth assuming a certain loss of generality to boost accuracy on the specific surroundings of the deployment. Specifically, results have shown that the edge model (MobileNetV2 with an SSD head for box regression) specialized for the context of a camera achieved between 2\% and 87\% higher accuracy than the same model trained using images from various cameras.
At the same time, results have shown how, using COVA, the specialization of the edge model for the context of specific cameras results in an increase of the \textit{off-the-shelf} accuracy by an average of 21\%, while keeping the computational cost constant, all through an automated process that uses a pre-trained model with state-of-the-art accuracy as teacher. However, the same results also highlight that special care should be taken in scenes where there is an imbalance on the class distribution of the objects captured. In such cases, reusing examples captured by cameras in similar contexts can help mitigate the issue.

It is worth stating that the edge model used throughout the experiments, even if arguably small, accounts for the non-negligible amount of 4.5 billion trainable parameters. It uses a MobileNetV2 architecture that can achieve a considerable 72\% accuracy on the notoriously generalist ImageNet and 19\% mAP on COCO when coupled with a Single Shot Detector head for box regression. Such neural network architectures can be useful, especially if they have been pre-trained on a related problem (even if it is loosely related). They offer a reasonable level of generalization that makes it easier to adapt to new environments, all at a relatively low computational cost. 
However, we believe there is a margin to further specialize the models used while keeping the same accuracy. Nevertheless, it is important to highlight the challenges involved in specializing neural networks. The more specialized a model is, the harder it is to use it reliably under uncontrolled environments because we are effectively increasing the number of possible inputs for which the model cannot generalize. Analogously, the better the automated analysis of a camera's context, the better and faster we can detect a concept drift, i.e., detect whenever the scene's context turns into something the model is not trained to recognize.

Consequently, as future work, we intend to explore alternative methods to further exploit the context of a camera. The modular design of COVA and its components open the door to seamlessly introduce new contextual features into the analysis of the scene, during the stages of automatic annotation and generation of the training dataset. We expect these new features to drive the generation of inexpensive and highly-specialized models that can still achieve high accuracy. Towards this goal, we intend to explore the use of custom and highly specialized neural networks for image classification and rely on the same techniques already used for motion detection to leverage localization. These networks are orders of magnitude smaller than the one used as the edge model previously mentioned. As a result, they are unable to generalize even in the slightest. In contrast, several of these can be trained for each camera context due to their small size and be quickly re-trained as soon as a concept drift is detected.

\begin{figure}
\centering
\begin{subfigure}[b]{\linewidth}
    \label{subfig:bcn-background}
    \includegraphics[width=\linewidth]{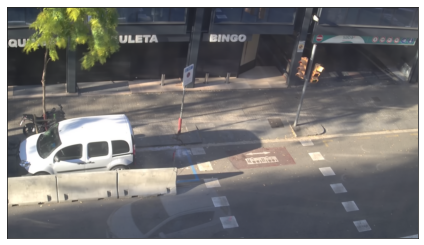}
    \caption{}
\end{subfigure}

\begin{subfigure}[b]{\linewidth}
    \includegraphics[width=\linewidth]{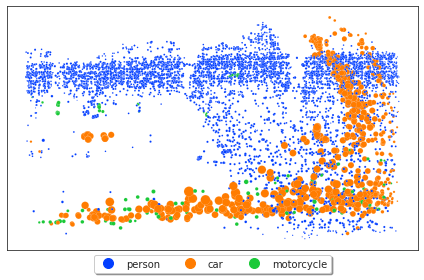}
    \caption{}
    \label{subfig:bcn-scatter}
\end{subfigure}
\caption{(a) Image captured by a static camera on a street. (b) Position (within the scene) and size of the objects captured by the camera throughout a day. Coordinates represent the centroid of the object and the point size is relative to the object size as per its detected bounding box.}
\label{fig:bcn-context} 
\end{figure}

We could argue that COVA leverages the context of a camera by exploiting the implicit inductive biases of convolutional neural networks, i.e., CNNs learn to consider the surroundings of an object and COVA uses during training images where objects are seen in the same context they will be when captured during inference. However, the context can offer much more than just visual information.
For example, when considering static cameras within a given context, objects tend to follow pre-established paths unless an extraordinary event occurs. From the point of view of a street camera, cars drive following the driveway, while pedestrians walk mostly on the sides of the street and the crosswalk. 
Therefore, the coordinates of an object can be indicative of the type of object.
Additionally, if the camera is placed in the middle of the street, cars and pedestrians get bigger or smaller as they get closer or further from the camera while following their paths. Consequently, the size of an object, along with its coordinates, can tell us its distance from the camera and, thus, be indicative of the type of object.
Figure~\ref{fig:bcn-context} depicts an example of how the type of object is strongly related to its position and size within the scene where it is captured. In the example, pedestrians (blue) walk mainly on the sidewalk and the crosswalk. Similarly, cars (orange) and motorcycles (green) are mostly captured on the driveway or entering the parking lot. Pedestrians and cars can be seen together in the region of the crosswalk, but the fact that not many people are the size of a car helps discern between them. Therefore, we observe a strong relationship between the position and size of an object with the type of object. It is important to highlight that the distribution shown in Figure~\ref{subfig:bcn-scatter} will vary even if the camera is only shifted in place, let alone moved to a different location. Therefore, these features can be considered part of a camera's context, and context is tightly coupled to its camera. It is precisely for this reason that the process of model specialization must be automated or, otherwise, it becomes unfeasible at scale.

In closing, we consider COVA to be the first step towards fully automated large-scale deployments for real-time edge video analytics.

%% file: main.bbl
\begin{thebibliography}{10}
\expandafter\ifx\csname url\endcsname\relax
  \def\url#1{\texttt{#1}}\fi
\expandafter\ifx\csname urlprefix\endcsname\relax\def\urlprefix{URL }\fi
\expandafter\ifx\csname href\endcsname\relax
  \def\href#1#2{#2} \def\path#1{#1}\fi

\bibitem{rivas2021performance}
D.~Rivas, F.~Guim, J.~Polo, D.~Carrera, Performance characterization of video
  analytics workloads in heterogeneous edge infrastructures, Concurrency and
  Computation: Practice and Experience (2021) e6317.

\bibitem{zhou2019edge}
Z.~Zhou, X.~Chen, E.~Li, L.~Zeng, K.~Luo, J.~Zhang, Edge intelligence: Paving
  the last mile of artificial intelligence with edge computing, Proceedings of
  the IEEE 107~(8) (2019) 1738--1762.

\bibitem{ananthanarayanan2017real}
G.~Ananthanarayanan, P.~Bahl, P.~Bod{\'\i}k, K.~Chintalapudi, M.~Philipose,
  L.~Ravindranath, S.~Sinha, Real-time video analytics: The killer app for edge
  computing, computer 50~(10) (2017) 58--67.

\bibitem{bilal2017edge}
K.~Bilal, A.~Erbad, {Edge computing for interactive media and video streaming},
  in: 2017 Second International Conference on Fog and Mobile Edge Computing
  (FMEC), IEEE, 2017, pp. 68--73.

\bibitem{huang2017speed}
J.~Huang, V.~Rathod, C.~Sun, M.~Zhu, A.~Korattikara, A.~Fathi, I.~Fischer,
  Z.~Wojna, Y.~Song, S.~Guadarrama, et~al., Speed/accuracy trade-offs for
  modern convolutional object detectors, in: Proceedings of the IEEE conference
  on computer vision and pattern recognition, 2017, pp. 7310--7311.

\bibitem{repo:tf_modelzoo}
{TensorFlow}, {TensorFlow 2 Detection Model Zoo},
  \url{https://github.com/tensorflow/models/blob/master/research/object_detection/g3doc/tf2_detection_zoo.md},
  [Accessed: 2021-11-10].

\bibitem{imagenet}
{ImageNet Large Scale Visual Recognition Challenge (ILSVRC)},
  \url{http://image-net.org/challenges/LSVRC/}, [Accessed: 2021-11-10].

\bibitem{bianco2018benchmark}
S.~Bianco, R.~Cadene, L.~Celona, P.~Napoletano, Benchmark analysis of
  representative deep neural network architectures, IEEE Access 6 (2018)
  64270--64277.

\bibitem{lin2014microsoft}
T.-Y. Lin, M.~Maire, S.~Belongie, J.~Hays, P.~Perona, D.~Ramanan,
  P.~Doll{\'a}r, C.~L. Zitnick, Microsoft coco: Common objects in context, in:
  European conference on computer vision, Springer, 2014, pp. 740--755.

\bibitem{tensorflow}
{TensorFlow}, \url{https://www.tensorflow.org/}, [Accessed: 2021-11-10].

\bibitem{openvino}
{OpenVINO Documentation}, \url{https://docs.openvino.ai/latest/index.html},
  [Accessed: 2021-11-10].

\bibitem{edge_repo}
{COVA - repository}, \url{https://github.com/HiEST/cova-tuner}, [Accessed:
  2021-11-20].

\bibitem{shi2016edge}
W.~Shi, J.~Cao, Q.~Zhang, Y.~Li, L.~Xu, {Edge computing: Vision and
  challenges}, IEEE Internet of Things Journal 3~(5) (2016) 637--646.

\bibitem{kang2019challenges}
D.~Kang, P.~Bailis, M.~Zaharia, {Challenges and Opportunities in DNN-Based
  Video Analytics: A Demonstration of the BlazeIt Video Query Engine.}, in:
  CIDR, 2019.

\bibitem{ali2018edge}
M.~Ali, A.~Anjum, M.~U. Yaseen, A.~R. Zamani, D.~Balouek-Thomert, O.~Rana,
  M.~Parashar, Edge enhanced deep learning system for large-scale video stream
  analytics, in: 2018 IEEE 2nd International Conference on Fog and Edge
  Computing (ICFEC), IEEE, 2018, pp. 1--10.

\bibitem{canel2019scaling}
C.~Canel, T.~Kim, G.~Zhou, C.~Li, H.~Lim, D.~G. Andersen, M.~Kaminsky, S.~R.
  Dulloor, {Scaling video analytics on constrained edge nodes}, arXiv preprint
  arXiv:1905.13536.

\bibitem{shen2017fast}
H.~Shen, S.~Han, M.~Philipose, A.~Krishnamurthy, Fast video classification via
  adaptive cascading of deep models, in: Proceedings of the IEEE conference on
  computer vision and pattern recognition, 2017, pp. 3646--3654.

\bibitem{cai2019once}
H.~Cai, C.~Gan, T.~Wang, Z.~Zhang, S.~Han, Once-for-all: Train one network and
  specialize it for efficient deployment, arXiv preprint arXiv:1908.09791.

\bibitem{kang2017noscope}
D.~Kang, J.~Emmons, F.~Abuzaid, P.~Bailis, M.~Zaharia, Noscope: optimizing
  neural network queries over video at scale, arXiv preprint arXiv:1703.02529.

\bibitem{hsieh2018focus}
K.~Hsieh, G.~Ananthanarayanan, P.~Bodik, S.~Venkataraman, P.~Bahl,
  M.~Philipose, P.~B. Gibbons, O.~Mutlu, {Focus: Querying large video datasets
  with low latency and low cost}, in: 13th $\{$USENIX$\}$ Symposium on
  Operating Systems Design and Implementation ($\{$OSDI$\}$ 18), 2018, pp.
  269--286.

\bibitem{kang2019blazeit}
D.~Kang, P.~Bailis, M.~Zaharia, Blazeit: Optimizing declarative aggregation and
  limit queries for neural network-based video analytics (2019).
\newblock \href {http://arxiv.org/abs/1805.01046} {\path{arXiv:1805.01046}}.

\bibitem{hung2018videoedge}
C.-C. Hung, G.~Ananthanarayanan, P.~Bodik, L.~Golubchik, M.~Yu, P.~Bahl,
  M.~Philipose, Videoedge: Processing camera streams using hierarchical
  clusters, in: 2018 IEEE/ACM Symposium on Edge Computing (SEC), IEEE, 2018,
  pp. 115--131.

\bibitem{gou2021knowledge}
J.~Gou, B.~Yu, S.~J. Maybank, D.~Tao, Knowledge distillation: A survey,
  International Journal of Computer Vision 129~(6) (2021) 1789--1819.

\bibitem{polino2018model}
A.~Polino, R.~Pascanu, D.~Alistarh, Model compression via distillation and
  quantization, arXiv preprint arXiv:1802.05668.

\bibitem{mullapudi2019online}
R.~T. Mullapudi, S.~Chen, K.~Zhang, D.~Ramanan, K.~Fatahalian, Online model
  distillation for efficient video inference, in: Proceedings of the IEEE/CVF
  International Conference on Computer Vision, 2019, pp. 3573--3582.

\bibitem{goyal2021selfsupervised}
P.~Goyal, M.~Caron, B.~Lefaudeux, M.~Xu, P.~Wang, V.~Pai, M.~Singh,
  V.~Liptchinsky, I.~Misra, A.~Joulin, P.~Bojanowski, Self-supervised
  pretraining of visual features in the wild (2021).
\newblock \href {http://arxiv.org/abs/2103.01988} {\path{arXiv:2103.01988}}.

\bibitem{purushwalkam2020demystifying}
S.~Purushwalkam, A.~Gupta, Demystifying contrastive self-supervised learning:
  Invariances, augmentations and dataset biases (2020).
\newblock \href {http://arxiv.org/abs/2007.13916} {\path{arXiv:2007.13916}}.

\bibitem{wang2018fully}
S.~Wang, Y.~Zhou, J.~Yan, Z.~Deng, Fully motion-aware network for video object
  detection, in: Proceedings of the European Conference on Computer Vision
  (ECCV), 2018, pp. 542--557.

\bibitem{beery2020context}
S.~Beery, G.~Wu, V.~Rathod, R.~Votel, J.~Huang, Context r-cnn: Long term
  temporal context for per-camera object detection, in: Proceedings of the
  IEEE/CVF Conference on Computer Vision and Pattern Recognition, 2020, pp.
  13075--13085.

\bibitem{beery2018recognition}
S.~Beery, G.~Van~Horn, P.~Perona, Recognition in terra incognita, in:
  Proceedings of the European Conference on Computer Vision (ECCV), 2018, pp.
  456--473.

\bibitem{daudt2018fully}
R.~C. Daudt, B.~Le~Saux, A.~Boulch, Fully convolutional siamese networks for
  change detection, in: 2018 25th IEEE International Conference on Image
  Processing (ICIP), IEEE, 2018, pp. 4063--4067.

\bibitem{benezeth2010comparative}
Y.~Benezeth, P.-M. Jodoin, B.~Emile, H.~Laurent, C.~Rosenberger, Comparative
  study of background subtraction algorithms, Journal of Electronic Imaging
  19~(3) (2010) 033003.

\bibitem{bouwmans2019deep}
T.~Bouwmans, S.~Javed, M.~Sultana, S.~K. Jung, Deep neural network concepts for
  background subtraction: A systematic review and comparative evaluation,
  Neural Networks 117 (2019) 8--66.

\bibitem{rivas2021large}
D.~Rivas, F.~Guim, J.~Polo, D.~Carrera, Large-scale video analytics through
  object-level consolidation, arXiv preprint arXiv:2111.15451.

\bibitem{hinton2015distilling}
G.~Hinton, O.~Vinyals, J.~Dean, Distilling the knowledge in a neural network,
  arXiv preprint arXiv:1503.02531.

\bibitem{dries2009adaptive}
A.~Dries, U.~R{\"u}ckert, Adaptive concept drift detection, Statistical
  Analysis and Data Mining: The ASA Data Science Journal 2~(5-6) (2009)
  311--327.

\bibitem{wang2017idk}
X.~Wang, Y.~Luo, D.~Crankshaw, A.~Tumanov, F.~Yu, J.~E. Gonzalez, Idk cascades:
  Fast deep learning by learning not to overthink, arXiv preprint
  arXiv:1706.00885.

\bibitem{french1999catastrophic}
R.~M. French, Catastrophic forgetting in connectionist networks, Trends in
  cognitive sciences 3~(4) (1999) 128--135.

\bibitem{li2019rilod}
D.~Li, S.~Tasci, S.~Ghosh, J.~Zhu, J.~Zhang, L.~Heck, Rilod: near real-time
  incremental learning for object detection at the edge, in: Proceedings of the
  4th ACM/IEEE Symposium on Edge Computing, 2019, pp. 113--126.

\bibitem{kirkpatrick2017overcoming}
J.~Kirkpatrick, R.~Pascanu, N.~Rabinowitz, J.~Veness, G.~Desjardins, A.~A.
  Rusu, K.~Milan, J.~Quan, T.~Ramalho, A.~Grabska-Barwinska, et~al., Overcoming
  catastrophic forgetting in neural networks, Proceedings of the national
  academy of sciences 114~(13) (2017) 3521--3526.

\bibitem{shmelkov2017incremental}
K.~Shmelkov, C.~Schmid, K.~Alahari, Incremental learning of object detectors
  without catastrophic forgetting, in: Proceedings of the IEEE International
  Conference on Computer Vision, 2017, pp. 3400--3409.

\bibitem{stauffer1999adaptive}
C.~Stauffer, W.~E.~L. Grimson, Adaptive background mixture models for real-time
  tracking, in: Proceedings. 1999 IEEE computer society conference on computer
  vision and pattern recognition (Cat. No PR00149), Vol.~2, IEEE, 1999, pp.
  246--252.

\bibitem{opencv_background_subtraction}
Opencv: Tutorial background subtraction,
  \url{https://docs.opencv.org/master/d1/dc5/tutorial_background_subtraction.html},
  [Accessed: 2021-11-10].

\bibitem{oh2011large}
S.~Oh, A.~Hoogs, A.~Perera, N.~Cuntoor, C.-C. Chen, J.~T. Lee, S.~Mukherjee,
  J.~Aggarwal, H.~Lee, L.~Davis, et~al., A large-scale benchmark dataset for
  event recognition in surveillance video, in: CVPR 2011, IEEE, 2011, pp.
  3153--3160.

\bibitem{everingham2010pascal}
M.~Everingham, L.~Van~Gool, C.~K. Williams, J.~Winn, A.~Zisserman, The pascal
  visual object classes (voc) challenge, International journal of computer
  vision 88~(2) (2010) 303--338.

\bibitem{repo:edge_model}
{TensorFlow}, {SSD MobileNet V2 320x320 Pre-Trained Model},
  \url{http://download.tensorflow.org/models/object_detection/tf2/20200711/ssd_mobilenet_v2_320x320_coco17_tpu-8.tar.gz},
  [Accessed: 2021-11-10].

\bibitem{repo:ref_model}
{TensorFlow}, {EfficientDet D2 768x768 Pre-Trained Model},
  \url{http://download.tensorflow.org/models/object_detection/tf2/20200711/efficientdet_d2_coco17_tpu-32.tar.gz},
  [Accessed: 2021-11-10].

\end{thebibliography}
